%% file: sample-manuscript.tex
\documentclass[acmsmall,screen]{acmart}
\AtBeginDocument{%
  }
\setcopyright{acmlicensed}
\copyrightyear{2026}
\acmYear{2026}
\acmDOI{XXXXXXX.XXXXXXX}
\acmJournal{CSUR}
\acmISBN{978-1-4503-XXXX-X/2018/06}

\usepackage{xspace}
\newcommand{\etal}{\emph{et al.}\xspace}

\begin{document}

\title{Modality-Aware Feature Matching in Visual and Vision-Language Applications: A Comprehensive Survey}

\author{Weide Liu}
\orcid{0000-0002-9855-4479}
\email{weide001@e.ntu.edu.sg}
\affiliation{%
  \institution{School of Computing and Artificial Intelligence, Jiangxi University of Finance and Economics}
  \city{Nanchang}
  \country{China}
}
\affiliation{%
  \institution{College of Computing and Data Science, Nanyang Technological University}
  \city{Singapore}
  \country{Singapore}
}

\author{Wei Zhou}
\orcid{0000-0003-3641-1429}
\email{zhouw26@cardiff.ac.uk}
\affiliation{%
  \institution{School of Computer Science and Informatics, Cardiff University}
  \city{Cardiff}
  \country{UK}
}

\author{Jun Liu}
\orcid{0000-0002-4365-4165}
\email{j.liu81@lancaster.ac.uk}
\affiliation{%
  \institution{School of Computing and Communications, Lancaster University}
  \city{Lancaster}
  \country{UK}
}

\author{Ping Hu}
\orcid{0000-0002-3121-9852}
\authornote{Corresponding Author}
\email{chinahuping@gmail.com}
\affiliation{%
  \institution{School of Computer Science and Engineering, University of Electronic Science and Technology of China}
  \city{Chengdu}
  \country{China}
}

\author{Jun Cheng}
\orcid{0000-0003-1786-6188}
\email{Cheng\_Jun@a-star.edu.sg}
\affiliation{%
  \institution{Institute for Infocomm Research, Agency for Science, Technology and Research (A*STAR)
}
  \city{Singapore}
  \country{Singapore}
}

\author{Jungong Han}
\orcid{0000-0003-4361-956X}
\email{jghan@tsinghua.edu.cn}
\affiliation{%
  \institution{Department of Automation, Tsinghua University
}
  \city{Beijing}
  \country{China}
}

\author{Weisi Lin}
\orcid{0000-0001-9866-1947}
\email{WSLin@ntu.edu.sg}
\affiliation{%
  \institution{College of Computing and Data Science, Nanyang Technological University}
  \city{Singapore}
  \country{Singapore}
}

\renewcommand{\shortauthors}{Liu et al.}

\begin{abstract}

Feature matching is a cornerstone task in computer vision, essential for applications such as image retrieval, stereo matching, 3D reconstruction, and SLAM. This survey comprehensively reviews modality-aware feature matching, exploring traditional handcrafted methods and emphasizing contemporary deep learning approaches across various modalities, including RGB images, depth images, 3D point clouds, LiDAR scans, medical images, and vision-language interactions. Traditional methods, leveraging detectors like Harris corners and descriptors such as SIFT and ORB, demonstrate robustness under moderate intra-modality variations but struggle with significant modality gaps. Contemporary deep learning-based methods, exemplified by learned detector-descriptors such as CNN-based SuperPoint and detector-free matchers such as transformer-based LoFTR, substantially improve robustness and adaptability across modalities.
We highlight modality-aware advancements, such as geometric and depth-specific descriptors for depth images, sparse and dense learning methods for 3D point clouds, attention-enhanced neural networks for LiDAR scans, and specialized solutions like the MIND descriptor for complex medical image matching. Cross-modal applications, particularly in medical image registration and vision-language tasks, underscore the evolution of feature matching to handle increasingly diverse data interactions. 

\end{abstract}

\ccsdesc[500]{Computing methodologies~Artificial intelligence}
\ccsdesc[500]{Computing methodologies~Computer vision}
\ccsdesc[500]{Computing methodologies~Matching}

\keywords{Feature-Matching, Single-Modality, Cross-Modality, Vision-Language, Medical Image}

\maketitle

\section{Introduction}
Feature matching is a fundamental task in computer vision, essential for various critical applications, including image retrieval, stereo matching, 3D reconstruction, and simultaneous localization and mapping (SLAM). To provide a systematic and comprehensive analysis of this domain, this survey organizes feature matching methods according to specific data modalities, covering both single-modality (RGB images, 3D data, medical images) and cross-modality (medical imaging and vision-language) scenarios.

We first introduce single-modality feature matching methods for RGB images and 3D data, highlighting the progression from early handcrafted detectors and descriptors to modern deep-learning-based approaches. Subsequently, we dive into medical imaging, a domain uniquely positioned at the intersection of single- and cross-modality matching, discussing the specialized approaches required for effective image registration across different medical modalities. We then address vision-language feature matching, emphasizing cross-modal alignment techniques that bridge visual and textual data. Finally, we outline potential future research directions, reflecting emerging trends and promising avenues in multi-modal and generalized feature matching.

Throughout this survey, we distinguish between two related paradigms of feature matching. Sections~2--4 primarily focus on geometric correspondence, where the objective is to estimate sparse or dense spatial matches between pixels, keypoints, or 3D points across views to support downstream tasks such as stereo matching and registration (e.g., semi-dense 2D correspondence estimation in Efficient LoFTR~\cite{wang2024efficient} and rotation-invariant 3D point cloud matching in RoITr~\cite{yu2023rotation}). In contrast, Section~5 focuses on semantic cross-modal alignment between visual content and language, where the goal is to associate an image or image region with a textual concept, question, or instruction in a shared embedding space (e.g., scalable image--text retrieval adaptation via Multiway-Adapter~\cite{Long2024MultiwayAdapter} and instruction-tuned vision--language modeling in InstructBLIP~\cite{dai2023instructblip}). While these paradigms differ in their supervision signals and evaluation protocols, they rely on many of the same principles; such as contrastive learning, metric learning, and cross-attention.

In the context of RGB image matching, techniques have evolved significantly from early handcrafted methods, including corner detectors such as the Harris operator \cite{Harris1988}, robust local descriptors like SIFT \cite{Lowe2004} and SURF \cite{Bay2006}, to efficient binary descriptors exemplified by ORB \cite{Rublee2011}.
While these model-driven approaches proved effective for intra-modality matching under moderate viewpoint and illumination changes, they often struggle with the larger domain gaps and sensing differences encountered across modalities.
Recent learning-based methods have therefore proposed to overcome these limitations: for example, SuperPoint \cite{Superpoint} employs a self-supervised CNN detector-descriptor trained on synthetic data to achieve robust feature correspondence.   LoFTR \cite{Sun2021} advances the field further with its transformer-based matching architecture, eliminating the need for explicit keypoint detection.

In the domain of 3D data, which includes RGB-D, LiDAR point clouds, 3D meshes, and multi-view 2D to 3D point sets, the feature matching techniques initially relied on geometric descriptors, such as Spin Images \cite{Johnson1999} and Fast Point Feature Histograms (FPFH) \cite{Rusu2009}, specifically designed to address rigid transformations and sparse data structures. Recent developments have increasingly used deep learning approaches, with methods such as 3DMatch \cite{Zeng2017}, FCGF \cite{Choy2019}, D3Feat \cite{Bai2020}, and transformer-based architectures such as Predator \cite{Huang2021}, significantly improving matching accuracy and robustness.

In medical imaging, feature matching often requires specialized strategies due to inherent intensity variations and anatomical deformations across different imaging modalities (e.g., MRI, CT, PET, ultrasound). Traditional methods such as mutual information (MI)~\cite{viola1997alignment} and normalized mutual information (NMI)~\cite{studholme1999nmi} laid the groundwork for multi-modal registration. Recent deep learning-driven approaches, including VoxelMorph \cite{balakrishnan2019voxelmorph} and DiffuseMorph \cite{kim2022diffusemorph}, integrate powerful unsupervised learning strategies, effectively handling deformable and intensity-based matching challenges.

For Vision-language feature matching, which integrates visual and textual information, driving tasks such as image captioning \cite{showtell15}, visual question answering \cite{vqa15}, and cross-modal retrieval \cite{karpathy15}. Key developments include contrastively trained dual-encoder models like CLIP \cite{Radford2021} and ALIGN \cite{Jia2021}, which enable scalable open-vocabulary retrieval and classification. Visual grounding methods, exemplified by transformer-based models such as MDETR \cite{Kamath2021} and GLIP \cite{Li2022}, have improved precise alignment between language and specific image regions. Additionally, open-vocabulary approaches have extended classification, detection, and segmentation beyond training labels, leveraging semantic embeddings from large-scale pre-training \cite{li2022language,Zhong2022,ghiasi2022scaling}. Ongoing challenges include compositional reasoning, robustness, bias mitigation, and evaluation scalability \cite{johnson2017clevr,bahdanau2019closure,zhao2021racial}, guiding future research toward more interactive, embodied, and continuously learning vision-language systems.

Compared to existing surveys, such as Xu et al.~\cite{xu2024local} (Information Fusion, 2024) focusing on detector-based vs.\ detector-free paradigms, Huang et al.~\cite{huang2024survey} (IET Image Processing, 2024) summarizing traditional vs.\ deep learning pipelines, and Ma~\cite{ma2021image} (IJCV, 2021) contrasting classical vs.\ deep image matching, our survey differs in its \emph{analytical lens}: we adopt a \textbf{modality-aware perspective} that treats sensing modality as a first-class factor shaping what ``matchability'' means and which priors are exploitable.

    \begin{enumerate}
      \item \textbf{Modality-driven taxonomy.}
      We organize feature matching by \emph{data modality} (RGB, depth/RGB-D, LiDAR and 3D point clouds, medical imaging such as X-ray/CT/MRI, and vision--language), because each modality induces distinct invariances, noise characteristics, and structural cues (appearance vs.\ geometry vs.\ deformation). These modality-specific properties, in turn, drive different design choices in detectors, descriptors, and matchers.

      \item \textbf{Cross-modality as a unifying thread with modality gaps made explicit.}
      Beyond surveying methods within each modality, we explicitly connect single-modality matching to cross-modality settings (e.g., multi-modal medical registration and vision--language alignment), and analyze how modality gaps alter similarity definitions, supervision signals, and evaluation protocols.

      \item \textbf{Comparative synthesis and transferable insights.}
      We provide within- and across-modality comparisons that distill which principles transfer broadly (e.g., attention-based matching, contrastive objectives) and which require modality-specific modeling (e.g., geometric invariance in 3D/LiDAR, intensity-robust similarity in medical images, semantic alignment in vision--language).
    \end{enumerate}

\begin{figure}[t]
  \centering
    \includegraphics[width=1\linewidth]{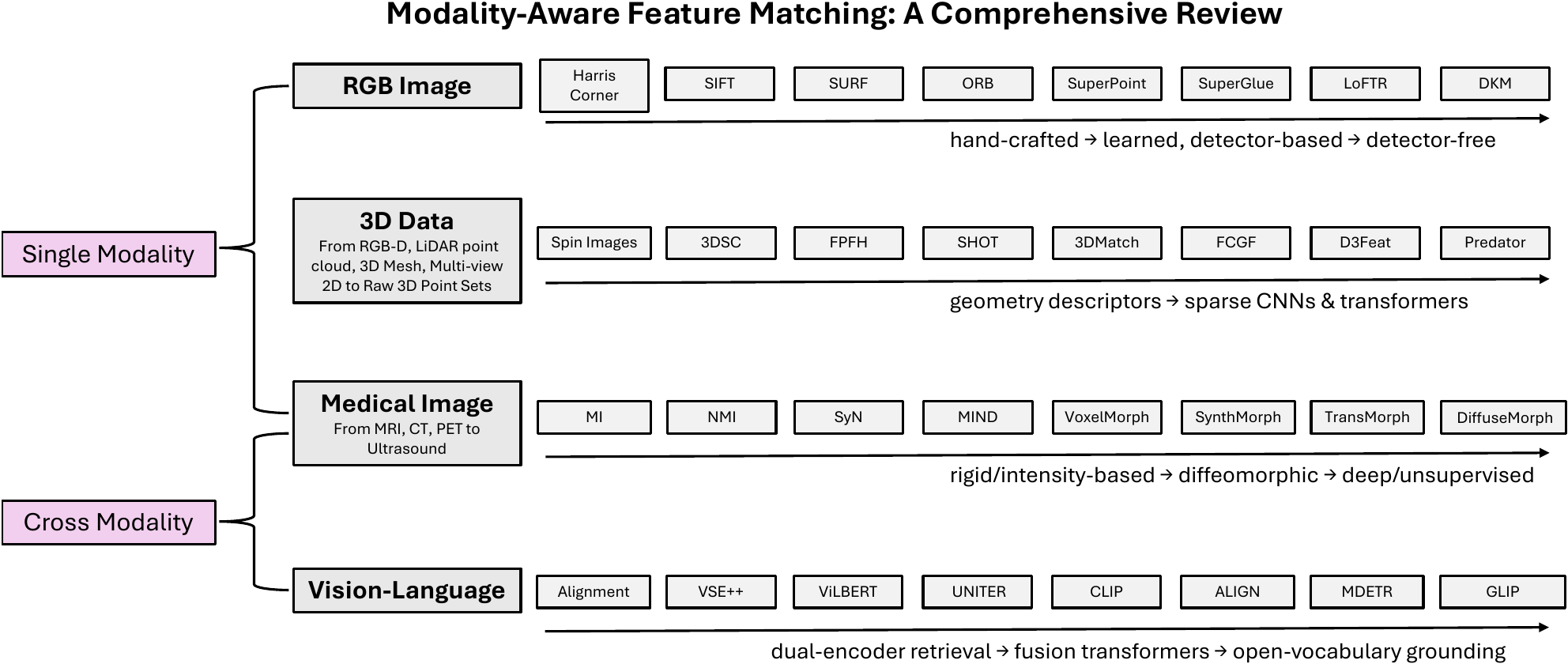}
    \caption{
An overview of modality-aware feature matching review, categorized by modality (RGB Image, 3D Data, Medical Image, and Vision-Language) and grouped into single-modality and cross-modality tasks. Each modality follows a chronological and methodological evolution. For RGB images, methods progress from classical handcrafted detectors (e.g., Harris Corner, SIFT, SURF) to learned sparse detectors (e.g., SuperPoint, SuperGlue) and detector-free dense matchers (e.g., LoFTR, DKM). In 3D data, the pipeline moves from geometric descriptors (e.g., Spin Images, FPFH) to sparse CNNs and transformer-based architectures (e.g., FCGF, Predator). Medical image registration techniques evolve from rigid/intensity-based methods (e.g., MI, SyN) to deep learning approaches (e.g., VoxelMorph, DiffuseMorph). Vision-language feature matching transitions from dual-encoder retrieval (e.g., VSE++) to fusion-based models (e.g., UNITER) and open-vocabulary grounding (e.g., MDETR, GLIP). The figure highlights the methodological shift in each domain, reflecting the growing reliance on deep learning and cross-modal understanding.
}
\vspace{-0.5cm}
    \label{Figure: Differnet_fusing}
\end{figure}

\begin{figure}[t]
  \centering
    \includegraphics[width=0.8\linewidth]{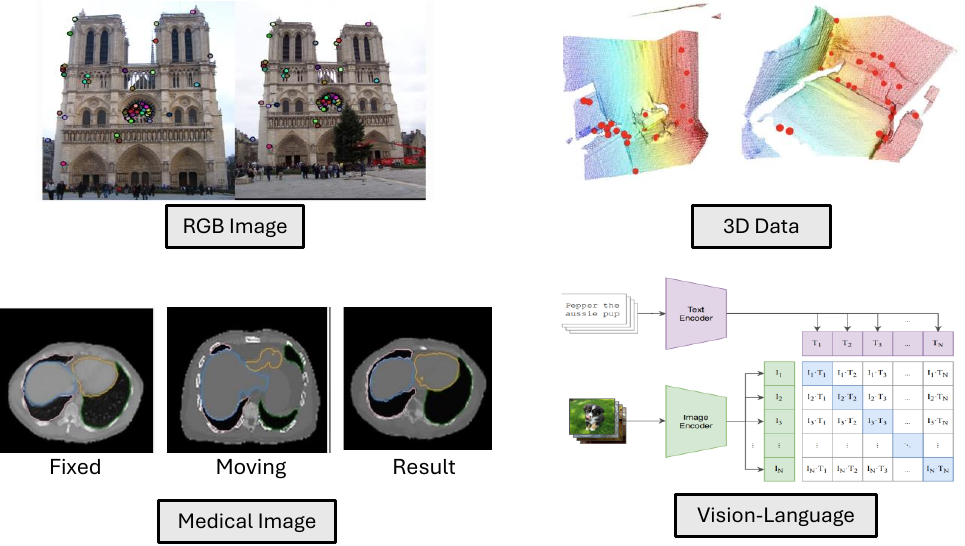}
    \caption{
Illustration of modality-aware feature matching.
The top row shows examples of single-modality matching. The RGB image pair demonstrates keypoint matching using SIFT~\cite{Lowe2004}. The 3D data shows local feature correspondences extracted by D3Feat~\cite{Bai2020}. The bottom row shows examples of cross-modality matching.
The left presents medical image registration results from TransMorph~\cite{chen2022transmorph} on an XCAT-to-CT alignment task, showing fixed, moving, and registered images. The bottom right visualizes vision-language alignment using CLIP~\cite{Radford2021}, which embeds text and image inputs into a shared representation space for contrastive learning.
}
\vspace{-0.5cm}
    \label{Figure: examples}
\end{figure}

In this survey, our discussion addresses unique challenges and methodologies related to single-modal feature matching (e.g., RGB, depth, medical imaging) and cross-modal scenarios (e.g., medical image registration, vision-language integration). We delineate the progression from classical detector-based pipelines towards contemporary detector-free solutions. 
Figure~\ref{Figure: Differnet_fusing} illustrates the overall pipeline of the survey, clearly mapping the evolution of feature matching methodologies across various data modalities. Furthermore, Figure~\ref{Figure: examples} offers representative examples of the results of the matching of modality-aware characteristics. In addition, we provide high-level visual summaries of the feature matching pipelines across different data modalities in the appendix for a convenient visual overview.

\input{2_Single_RGB}
\input{3_Depth_Image}
\input{4-5-combined}
\input{6_cross_vision_language}

\input{7_discussion_conclusion}

\input{8_Appendix}

\bibliographystyle{ACM-Reference-Format}
\bibliography{egbib}

\end{document}

%% file: 2_Single_RGB.tex
\section{RGB Images}\label{sec:rgb-features}
Feature matching in RGB images traditionally follows a pipeline of \textit{feature detection}, \textit{feature description}, and \textit{feature matching}. Over the decades, a lot of methods have been proposed for each stage of this pipeline, from handcrafted approaches to more recent deep learning-based techniques.

\subsection{Handcrafted Feature Detectors and Descriptors}
Early work on feature matching was dominated by handcrafted methods that rely on explicit formulas and heuristics for detection and description.  Notable corner detectors include the Harris detector \cite{Harris1988}, which finds points with significant intensity variation in two orthogonal directions by analyzing the autocorrelation matrix, and the Shi–Tomasi “Good Features to Track” criterion \cite{Shi1994} that selects corners based on the smaller eigenvalue of that matrix for improved tracking stability. These corner detectors assume a fixed scale; to achieve scale or affine invariance, researchers introduced multi-scale or affine adaptation. For instance, the Harris-Laplace and Hessian-Laplace detectors \cite{Mikolajczyk2005IJCV} measure over a Laplacian-of-Gaussian (LoG) scale space to detect scale-invariant points. Another influential detector is the Maximally Stable Extremal Region (MSER) detector \cite{Matas2002}, which finds blob-like regions by thresholding intensity and selecting extremal regions that are stable across thresholds; MSER yields affine-covariant regions and was widely used in wide-baseline matching.

Later, Lowe \cite{Lowe2004} proposed the Scale-Invariant Feature Transform (SIFT) for robust scale-invariant features. SIFT’s detector identifies blob-like keypoints in scale-space as local maxima of Difference-of-Gaussians. Numerous subsequent methods sought improvements in speed or invariance. For example, Speeded-Up Robust Features (SURF) \cite{Bay2006} employs an integer approximation of LoG using box filters for fast scale-space detection, and a descriptor based on Haar wavelet responses, reducing computation while maintaining robustness. Another family of fast detectors is those based on binary intensity tests, such as the FAST corner detector \cite{Rosten2006}. FAST classifies a pixel as a corner by examining a small circle of neighboring pixels.

Handcrafted descriptors have likewise evolved. The SIFT descriptor remained dominant for its robustness, but alternatives were proposed to reduce dimensionality or computation. PCA-SIFT \cite{Ke2004} showed that applying Principal Component Analysis to normalized gradient patches can compress SIFT’s descriptor to a shorter vector while preserving discrimination. Similarly, the Gradient Location-Orientation Histogram (GLOH) descriptor \cite{Mikolajczyk2005PAMI} extended SIFT by using a log-polar location grid and PCA compression to increase distinctiveness. Later, there was a shift toward binary descriptors for efficiency in matching. BRIEF \cite{Calonder2010} was an early binary descriptor: instead of histograms, it compares pixel intensities in a smoothed patch according to a predefined random pattern of point pairs, producing a bit-string descriptor. BRIEF is very fast to compute and match, but it is not rotation-invariant. ORB \cite{Rublee2011} builds on FAST detection and BRIEF description, adding an orientation assignment and learning a rotated binary test pattern for robustness to rotation. ORB became popular as a free alternative to SIFT, offering near real-time performance while retaining some invariance \cite{Rublee2011}. Other notable binary descriptors include BRISK \cite{Leutenegger2011}, which uses patterns of point pairs at multiple scales and short/long pairings to improve rotation and scale invariance, and FREAK \cite{Alahi2012}, which uses a retinal sampling pattern to achieve robustness with very few bits. 

Beyond corners and blobs, researchers also developed region detectors capturing larger structures. For example, edge-based and intensity-based region detectors \cite{Tuytelaars2004} identify invariant regions tied to image boundaries or intensity patterns. Kadir and Brady’s salient region detector \cite{Kadir2001} selects regions that maximize entropy across scales, which are informative regions for matching. Another distinctive approach is the SUSAN detector \cite{Smith1997}, which defines corners via the “Univalue Segment Assimilating Nucleus” principle, avoiding explicit gradient computation and instead counting contiguous areas of similar brightness around a pixel. 

To further improve reliability, robust estimation methods like RANSAC \cite{Fischler1981} are applied to the set of putative matches to filter out outliers by fitting a geometric model. Numerous variants of RANSAC have been proposed to improve its speed or accuracy. Notable examples include: PROSAC \cite{Chum2005}, which leverages an assumed ranking of matches by quality to prioritize sampling high-quality correspondences first, yielding faster convergence; LO-RANSAC \cite{Chum2003} adds a local optimization step to polish the model estimate after RANSAC finds a consensus, thereby increasing accuracy; GC-RANSAC \cite{Barath2018} introduces statistically robust model scoring (M-estimator sampling consensus) and Graph-Cut optimization to handle varying inlier scales and improve stability; and other specialized versions like GroupSAC \cite{Ni2009} for situations with grouped structures, or EVSAC \cite{Fragoso2013} models matching scores with extreme value theory to guide hypothesis generation. In addition, matching algorithms that enforce global consistency have been explored. For example, Leordeanu and Hebert’s spectral matching \cite{Leordeanu2005} casts feature matching as a graph correspondence problem and uses eigen-decomposition to find an assignment that maximizes pairwise alignment consistency. Similarly, reweighted random walk matching \cite{Cho2010} uses random walk probabilities on a correspondence graph to find a coherent set of matches. 

Overall, handcrafted methods have been extensively evaluated in literature. Classic benchmarks by Schmid \emph{et al.} \cite{Schmid2000} compared early corner detectors, introducing the repeatability metric (the percentage of true scene points re-detected under varying imaging conditions) as a standard for detector performance. Similarly, Mikolajczyk and Schmid’s influential study evaluated descriptors (SIFT, PCA-SIFT, shape context, steerable filters, etc.) under rotations, scale changes, and affine transformations, using recall-precision curves to measure distinctiveness \cite{Mikolajczyk2005PAMI}. Their findings confirmed SIFT’s superiority among early descriptors and spurred interest in developing new ones. The combination of multi-scale detectors with SIFT or SURF descriptors and robust matching via RANSAC became the dominant paradigm for feature matching in RGB images. Open-source implementations (e.g., OpenCV \cite{bradski2000opencv}) further popularized these methods. However, limitations remained: handcrafted features can fail under extreme appearance changes, repetitive patterns, or a lack of semantic robustness. This opened the door for learning-based approaches to further improve the resilience and specificity of feature matching.

\subsection{Deep Learning-Based Local Features}
The resurgence of neural networks in vision prompted researchers to replace or augment each stage of the feature matching pipeline with learned components. Early attempts focused on learning better descriptors for patches extracted by traditional detectors. For example, instead of using SIFT’s analytic descriptor, \textbf{CNN-based patch descriptors} were trained to discriminate between matching and non-matching image patches. One of the first such efforts compared features from convolutional neural networks pre-trained for ImageNet classification and found they could outperform SIFT on patch matching benchmarks \cite{Fischer2014}. Han \etal’s MatchNet \cite{Han2015} trained a Siamese network to directly predict if two patches match, effectively learning the descriptor and a similarity metric jointly. Zagoruyko and Komodakis \cite{Zagoruyko2015} explored various Siamese network architectures to learn patch similarity, introducing metric learning losses that significantly improved patch matching accuracy over SIFT. Simo-Serra \etal trained a DeepDesc \cite{SimoSerra2015} using a hinge loss to ensure true match patches are closer than non-matches, showing the benefit of larger training data and hard-negative mining in learning descriptors. Balntas \etal’s TFeat \cite{Balntas2016} simplified the architecture to a shallow CNN and used triplet loss with positive and negative patch examples to learn a compact descriptor efficiently. In the meanwhile, Tian \etal’s L2-Net \cite{Tian2017} demonstrated that with appropriate training, a fully convolutional descriptor could substantially outperform SIFT. L2-Net achieved state-of-the-art patch matching accuracy and introduced the widely used “hardest-in-batch” triplet loss \cite{Tian2017}. Shortly after, HardNet \cite{Mishchuk2017} further refined the training by directly optimizing the structured margin with the hardest negatives across the entire batch, producing a descriptor that consistently topped benchmarks and became a common baseline in learned descriptor research. Some works also explored learning binary descriptors to retain efficiency. For instance, BinBoost \cite{Trzcinski2015} used machine learning to optimize binary tests or projections for descriptor generation. Although deep networks for binary descriptors have been less common, a few methods like DOAP~\cite{He_2018_CVPR}, which learns binary descriptors via ordinal embedding, approached that, but real-valued learned descriptors remained dominant due to their higher accuracy.

While descriptors were being learned, researchers also tackled \textbf{learning better keypoint detectors}. One challenge is that keypoint detection involves a non-differentiable selection (argmax or non-max suppression), making it hard to train in an end-to-end manner. Early attempts sidestepped this by training to mimic an existing detector’s output under new conditions. For example, TILDE \cite{Verdie2015} learned a detector robust to illumination changes by training on sequences of images over time (e.g., day-night cycles). It regressed a heatmap of “detector responses” whose local maxima are stable interest points under severe lighting changes. Another landmark was LIFT \cite{Yi2016}, which presented the first fully learned pipeline: a deep network was trained end-to-end to perform detection, orientation assignment, and description. Shortly after, LF-Net \cite{Ono2018} improved on this by formulating the detector and descriptor as a unified differentiable architecture and using self-supervision to reduce the need for 3D ground truth. LF-Net’s detector was explicitly trained for repeatability and reliability, marking a step toward truly learnable detectors.

Another direction is joint learning of detection and description. D2-Net \cite{Dusmanu2019} took a bold approach: dispense with a separate detector entirely and use a deep CNN to generate a dense feature map from the image, then simply detect keypoints as local peaks in the feature maps. D2-Net showed strong results, especially in difficult conditions, and highlighted the advantage of dense, high-level features for locating useful keypoints beyond classical corner definitions. However, it also produced many keypoints on textured areas and had no built-in mechanism to suppress less reliable points. This was addressed by R2D2 \cite{Revaud2019}, which learned to predict not only dense descriptors but also a “reliability” score for each location. By selecting keypoints with both high repeatability and high reliability, R2D2 achieved a better trade-off of precision and recall. Meanwhile, other efforts like Luo \etal’s GeoDesc \cite{Luo2018} improved descriptors by incorporating geometric constraints and global context during training.

A parallel direction incorporated global context and learning-to-match strategies on top of local features. For example, ContextDesc \cite{Luo2019} augments local descriptors by incorporating cross-modality context: it combined visual context (from a higher-level CNN feature of the whole image) and geometric context (spatial distribution of nearby keypoints) with the local descriptor using a learned fusion \cite{Luo2019}. This improved matching in ambiguous regions by mimicking how humans consider their surroundings when matching points. The trend of using context culminated in SuperGlue \cite{Sarlin2020}, which is not a detector or descriptor but a learned matching network. SuperGlue takes as input two sets of keypoints and their descriptors (e.g., from SIFT or SuperPoint) and uses a Graph Neural Network with self- and cross-attention to directly predict an optimal matching between the two sets \cite{Sarlin2020}. By treating the problem as graph matching, SuperGlue can enforce one-to-one match constraints and leverage context from neighboring keypoints to resolve ambiguous matches. It significantly outperformed naive nearest-neighbor matching, essentially learning to be an “intelligent matcher” that performs mutual compatibility reasoning. SuperGlue set a new state-of-the-art in feature matching and has been used in downstream tasks like SLAM and SfM, effectively achieving an end-to-end learned matching pipeline when combined with a learned detector like SuperPoint~\cite{Superpoint}. A lighter successor, LightGlue \cite{Lindenberger2023LightGlue}, further optimized this approach to run faster with minimal loss of accuracy, making learned matching more practical.

Another major advance was the rise of transformer-based detector-free approaches. Transformers, with their strong ability to capture long-range dependencies via self-attention, have proven very effective in matching two images without requiring explicit keypoint detection. The pioneering work in this vein is LoFTR by Sun \etal \cite{Sun2021}, which establishes correspondences on a grid of image features by alternating self- and cross-attention between the two images’ feature maps. LoFTR produces dense or semi-dense matches and can handle low-texture areas by effectively “hallucinating” matches guided by global context \cite{Sun2021}. It first computes coarse matches on a downsampled image feature map and then refines matches at finer resolution, yielding highly accurate correspondences even without a traditional keypoint detector. LoFTR and subsequent transformer matchers bypass the need to detect keypoints beforehand, instead letting the network attend to all pixels and pick out matches. Variants and improvements soon followed: COTR (Correspondence Transformer) \cite{Jiang2021} uses a similar idea but with a different query mechanism for correspondence prediction; QuadTree Attention \cite{tang2022quadtree} and MatchFormer \cite{Wang2022} introduced hierarchical attention to improve efficiency for high-resolution images, reducing the quadratic cost of global attention while preserving accuracy. These detector-free methods, though computationally heavier, have achieved impressive results on difficult wide-baseline problems, finding matches in scenes with weak texture or repetitive patterns where sparse detectors struggle. Dense matching networks have also been adapted for feature matching; for example, NCNet \cite{Rocco2018} performs differentiable nearest-neighbor matching followed by a neighborhood consensus network to find coherent matches across images. Such methods treat matching as a learning problem globally, rather than focusing on local descriptor invariance alone.

In summary, deep learning-based methods have greatly advanced the state of RGB image matching. They have produced more repeatable detectors (learning where to detect points that survive viewpoint/illumination changes), more discriminative descriptors (learning how to describe patches for higher distinctiveness), and even entirely new paradigms for matching (learning to match with attention mechanisms, and end-to-end trainable pipelines). Table \ref{tab:top10features} highlights some of the influential methods from both the handcrafted and deep learning eras. These modern techniques, when evaluated on challenging benchmarks, significantly outperform classic methods in terms of matching precision and robustness, although often at a higher computational cost. Nonetheless, due to their task-specific optimizations and ability to learn from data, deep feature matching methods have become the go-to choice in applications requiring high reliability under difficult conditions (e.g., long-term visual localization, as evidenced by SuperGlue and LoFTR’s superior results in day-night matching challenges). Active research continues into making these networks faster, more generalizable, and easier to train, bridging the gap between handcrafted efficiency and deep learning performance. These dense correspondence principles also extend to stereo matching, where evidential fusion of local and global cues~\cite{lou2023elfnet}, together with explicit modeling of intra- and cross-view geometry~\cite{gong2024learning,zeng2025uncertainty}, yields more accurate disparity estimation.

\begin{table*}[ht]
\centering
\caption{The local feature matching methods for RGB images.}
\label{tab:top10features}
\resizebox{\linewidth}{!}{%
\begin{tabular}{p{3.3cm} p{5.8cm} p{6.2cm}}
\toprule
\textbf{Method} & \textbf{Brief Description} & \textbf{Key Advantage} \\ \midrule
Harris Corner \cite{Harris1988} & Detects corner points via the autocorrelation matrix; finds points with high intensity variation in orthogonal directions. & Very fast; high repeatability under small transformations; foundation for many later detectors. \\
SIFT \cite{Lowe2004} & Scale‐Invariant Feature Transform: DoG blob detector plus 128-dimensional gradient‐histogram descriptor. & Invariant to scale and rotation; robust to viewpoint and lighting changes; highly distinctive descriptors. \\
SURF \cite{Bay2006} & Blob detector/descriptor using Haar wavelets and integral images (approximate DoG). & Faster than SIFT; some affine invariance; good speed/robustness balance. \\
FAST \cite{Rosten2006} & Corner detector using machine-learned pixel‐intensity tests on a circle around each candidate pixel. & Extremely fast (real-time); widely used in SLAM and ORB pipelines. \\
ORB \cite{Rublee2011} & Oriented FAST + Rotated BRIEF; binary descriptor on multi-scale FAST keypoints. & Very fast matching; patent-free; ideal for mobile/AR real-time use. \\
LIFT \cite{Yi2016} & End-to-end CNN detector, orientation estimator, and descriptor trained with SfM supervision. & First fully learned local feature pipeline; better under challenging conditions than SIFT. \\
SuperPoint \cite{Superpoint} & Self-supervised CNN detector/descriptor. & Single forward pass for thousands of keypoints; no manual labels; excellent speed/quality trade-off. \\
D2-Net \cite{Dusmanu2019} & Dense CNN features: keypoints are peaks in feature maps, descriptors from the same network. & Joint detection/description; reliable in low-texture areas; strong invariance from deep features. \\
SuperGlue \cite{Sarlin2020} & Graph neural-network matcher refining SuperPoint (or other) features with attention. & Learns globally consistent correspondences; greatly improves precision/recall. \\
LoFTR \cite{Sun2021} & Detector-free transformer matcher producing coarse-to-fine dense matches. & Handles extreme viewpoint changes; dense matches without explicit keypoints. \\ 
DKM \cite{Edstedt2023DKM} & Dense matching framework with kernelized global matcher, stacked depthwise warp refinement, and depth-based certainty estimation. & Robustly handles large viewpoint variations and challenging illumination conditions with accurate dense correspondences.\\
\bottomrule
\end{tabular}%
}
\end{table*}

\subsection{Benchmark Datasets and Evaluation Protocols}

Feature matching research has been guided by standardized benchmarks and evaluation metrics. An early classic is the Oxford affine covariant regions dataset~\cite{Mikolajczyk2005IJCV}, which provides image sequences of the same scene under known homographies simulating viewpoint and illumination changes (e.g., \textit{Graffiti}, \textit{Wall}, \textit{Bark}, \textit{Boat}). It established standard protocols for evaluating detectors via repeatability and descriptors via recall and precision–recall curves. Table~\ref{tab:matching-datasets} summarizes influential datasets and evaluation protocols, and Table~\ref{tab:rgb-hpatches} reports method performance on the HPatches dataset~\cite{Balntas2017}.

\begin{table*}[htbp]\small
\centering
\caption{The public academic datasets commonly used to evaluate single-modality RGB feature matching.}
\label{tab:matching-datasets}

\resizebox{\textwidth}{!}{%
\begin{tabular}{@{}p{2.8cm} p{2.8cm} p{5.1cm} p{5.1cm}@{}}
\toprule
\textbf{Dataset} & \textbf{Use-case} & \textbf{Description} & \textbf{Typical Evaluation Protocol} \\ \midrule

Oxford Affine \cite{Mikolajczyk2005IJCV} &
Planar scenes; viewpoint / illumination change &
40 image pairs (8 scenes, 5 transformations each) with known homographies. &
Repeatability and matching score for keypoint detectors / descriptors under viewpoint and lighting change. \\

UBC PhotoTour \cite{Brown2010} &
Patch correspondence (3-D scenes) &
$\sim$2.5 M $64{\times}64$ patch pairs from three SfM scenes (Liberty, Notre Dame, Yosemite) with ground-truth labels. &
Training / testing local descriptors via ROC curves, FPR@TPR, etc. \\

Wide-Baseline Stereo \cite{Tola2010} &
Wide-baseline stereo pairs &
47 pairs from two short sequences plus additional wide-baseline images with depth maps. &
Dense descriptor and stereo-matching evaluation under large baselines. \\

HPatches \cite{Balntas2017} &
Planar homography sequences &
116 six-image sequences with known homographies (59 viewpoints, 57 illuminations). &
Standard benchmark for detection / description; homography-estimation accuracy. \\

KITTI 2012/2015 \cite{Geiger2013} &
Driving stereo (outdoor) &
Urban stereo pairs ($\approx$200 training + 200 test) with LiDAR ground-truth disparity / optical flow. &
Matching accuracy and endpoint error in dynamic scenes. \\

Strecha MVS \cite{Strecha2008} &
Multi-view; wide baseline &
Outdoor multi-view stereo (e.g., Fountain-P11, Herz-Jesu-P8) with calibrated cameras and dense 3-D ground truth. &
3-D reprojection error for SfM / MVS robustness to strong viewpoint change. \\

ETH3D \cite{Schops2017} &
Multi-view (indoor / outdoor) &
High-resolution images with ground-truth poses and point clouds; 20 stereo pairs + several multi-view sets. &
Two-view matching and full SfM / MVS evaluation with precise geometry. \\

ScanNet \cite{Dai2017} &
Indoor RGB-D video &
1613 scans ($\approx$2.5 M frames) with depth, ground-truth trajectories, and surface reconstructions. &
SLAM / odometry tests: relocalization accuracy, frame-to-frame matching. \\

YFCC100M (landmarks) \cite{Thomee2016} &
Internet photo collections (SfM) &
100 M Flickr images; landmark subsets (e.g., 4,000 pairs) with SfM poses and depth. &
Large-scale relative-pose evaluation on crowdsourced imagery. \\

Aachen Day–Night \cite{Sattler2018} &
Day vs.\ night localization &
4,328 daytime references + 922 queries (824 day, 98 night) of urban scenes. &
Pose recall/inlier counts across extreme illumination change. \\

PhotoTourism (IMC 2020) \cite{Jin2020} &
Crowdsourced landmark SfM &
Large landmark collections used in the CVPR 2020 Image Matching Challenge; COLMAP poses and depth. &
Pose and reconstruction accuracy in wide-baseline landmark imagery. \\

ZEB (Zero-Shot Eval.) \cite{Shen2024} &
Cross-domain matching &
Benchmark of 46 k pairs from 12 datasets (8 real, 4 synthetic) spanning resolutions, scenes, and overlaps (10–50 \%). &
Generalization test across domains; reported via pose and inlier metrics. \\ \bottomrule
\end{tabular}%
}
\end{table*}

\begin{table}[htbp]
\small
\centering
\caption{Estimation accuracy on the HPatches dataset~\cite{Balntas2017}. }
\begin{tabular}{l|c|c|c}
\toprule
Method & AUC@3px & AUC@5px & AUC@10px \\
\midrule
SIFT \cite{Lowe2004} & 24.0 & 40.0 & 57.0 \\
ORB \cite{Rublee2011} & 20.5 & 35.7 & 50.2 \\
SuperPoint \cite{Superpoint} + SuperGlue \cite{Sarlin2020} & 53.9 & 68.3 & 81.7 \\
D2Net \cite{Dusmanu2019} & 23.2 & 35.9 & 53.6 \\
R2D2 \cite{Revaud2019} & 50.6 & 63.9 & 76.8 \\
LoFTR \cite{Sun2021} & 65.9 & 75.6 & 84.6 \\
QuadTree Attention \cite{tang2022quadtree} & 66.3 & 76.2 & 84.9 \\
ASpanFormer \cite{Chen2022_single} & 67.4 & 76.9 & 85.6 \\
DKM \cite{Edstedt2023DKM} & 71.3 & 80.6 & 88.5 \\
\bottomrule
\end{tabular}
\label{tab:rgb-hpatches}
\end{table}

%% file: 3_Depth_Image.tex
\section{3D   Data: From RGB-D, LiDAR point cloud, 3D Mesh to Multi-view 2D}\label{sec:depth_features}
3D images (such as those from RGB-D sensors or LiDAR range scans) provide geometric information that enables feature matching based on 3D shape cues rather than appearance. A rich body of work has been devoted to detecting keypoints and computing local descriptors directly on 3D data for tasks like point cloud registration, 3D object recognition, and SLAM. This section reviews both classical handcrafted approaches and deep learning-based methods for feature matching using 3D data. We also summarize common benchmark datasets and evaluation protocols, and highlight representative applications of 3D image feature matching. 

\subsection{Handcrafted Feature Detectors and Descriptors}
Early research on depth images adapted 2D feature concepts to 2.5D range data and 3D surfaces. One seminal work is the spin image descriptor \cite{Johnson1999}, which projects a 3D point’s neighborhood onto a 2D histogram using radial and elevation distances to form a rotation-invariant surface descriptor. Spin images demonstrated robust object recognition in cluttered scenes using only geometric shape information \cite{Johnson1999}. Another influential descriptor is the 3D Shape Context (3DSC) \cite{Frome2004}, which extends 2D shape context to 3D by histogramming the spatial distribution of neighboring points in a spherical grid around the keypoint.

Subsequent approaches focused on capturing local surface geometry with greater invariance and efficiency. Rusu \emph{et al.} proposed the Point Feature Histogram (PFH) \cite{Rusu2008}, which encodes the pairwise geometric relations (angles and distances) between points in a neighborhood. Its accelerated variant, Fast Point Feature Histogram (FPFH) \cite{Rusu2009}, reduces computation by considering only a keypoint’s direct neighbors for the histogram accumulation. 
To detect repeatable keypoints in range data, several techniques have adapted corner detectors to 3D surfaces. For example, the Harris 3D detector \cite{Sipiran2011} extends the Harris corner criterion to 3D mesh geometry, and the Intrinsic Shape Signature (ISS) keypoint \cite{Zhong2009} selects points with salient local variation based on eigenanalysis of the neighborhood’s covariance matrix (avoiding flat or linear regions).
Researchers also developed descriptors tailored specifically for organized range images (“2.5D” data). Lo \cite{Lo2009} introduced 2.5D SIFT, which applies SIFT-style filtering on depth images to obtain invariant keypoints and descriptors. Bayramoglu and Alatan proposed Shape Index SIFT (SI-SIFT), combining shape index maps of the depth image with SIFT to handle rotational and scale changes in range data \cite{bayramoglu2010shape}. These methods leveraged the structured nature of depth images (rows and columns corresponding to sensor view) to directly extend 2D feature detectors into the depth modality.

To achieve rotation invariance in 3D, many handcrafted descriptors compute a local reference frame (LRF) at each keypoint. The Unique Shape Context (USC) of Tombari \emph{et al.} is an LRF-aligned extension of 3D shape context that avoids multiple descriptors per point by consistently orienting the neighborhood \cite{Tombari2010}. Tombari \emph{et al.} also developed the SHOT descriptor (Signature of Histograms of Orientations) \cite{Tombari2010b}, which builds an LRF and then aggregates point counts weighted by normals into a set of angular histograms. Other notable descriptors include Spin Images variations (e.g., Tri-Spin Images integrating multiple spin image projections) \cite{Guo2014}, 3D SURF adaptations using integral images on depth data, and RoPS (Rotational Projection Statistics) by Guo \emph{et al.}, which projects the 3D neighborhood onto multiple planes and computes statistical moments to form a descriptor \cite{Guo2013}. RoPS, coupled with a robust LRF, was shown to be highly distinctive for object retrieval and recognition in 3D scenes, outperforming many earlier features under clutter and occlusions \cite{Guo2013}. Other notable descriptors include TOLDI \cite{Yang2017}, which generates a Triple Orthogonal Local Depth Image descriptor by projecting the local surface onto three orthogonal planes. 

However, these methods can be sensitive to noise, point density variations, and occlusions. Moreover, they often produce high-dimensional descriptors and require fine-tuned parameters for each dataset. Despite these limitations, handcrafted features laid the groundwork for depth image matching and are still employed in real-time systems and as benchmarks. They enable correspondence matching in applications like coarse object alignment and SLAM loop closure detection, where no training data are available.

\subsection{Deep Learning-Based Methods}
One challenge in learning 3D features is how to represent the input. Different strategies have been explored: 3DMatch used voxel grids with truncated distance function values as input \cite{Zeng2017}. Other works project local 3D surfaces into multi-view images: for example, Bai \emph{et al.} in D3Feat mention that early attempts like Zeng’s volumetric CNN and multi-view approaches were later complemented by fully point-based networks. Point-based methods, inspired by PointNet~\cite{qi2016pointnet}, feed raw point coordinates (or point pairs) into a network. PPFNet \cite{Deng2018a} is one such example: it augments PointNet with point-pair features (distances and angles) to incorporate geometric context, yielding a learned descriptor that is more rotation-robust than point-only input \cite{Deng2018a}. An unsupervised variant, PPF-FoldNet \cite{Deng2018b}, uses a folding-based autoencoder to learn descriptors without manual correspondences, achieving rotation invariance through on-the-fly data augmentation.

Another direction leverages fully-convolutional networks on sparse 3D data. Choy \emph{et al.} proposed FCGF (Fully Convolutional Geometric Features) \cite{Choy2019}, which uses a 3D sparse CNN (with Minkowski convolution) to compute a dense grid of descriptors over the space, rather than extracting per-patch descriptors. Despite its efficiency, it achieves state-of-the-art accuracy on indoor (3DMatch) and outdoor (KITTI) benchmarks. Bai \emph{et al.} extend this idea in D3Feat \cite{Bai2020}, which not only computes dense descriptors via a U-Net style sparse CNN but also learns to predict which points are good keypoints. By jointly training a detection and description head (with self-supervision for the detector using mutual consistency of matches), D3Feat finds repeatable keypoints and descriptors that yield improved registration results compared to using fixed detectors like ISS or random sampling \cite{Bai2020}. 

Rotation invariance is a crucial issue for 3D features – many early learned models were not inherently rotation-invariant and relied on data augmentation. SpinNet maps the local 3D neighborhood onto a cylindrical coordinate space aligned with the keypoint’s principal axes, and then applies 3D convolutions \cite{Ao2021}. This architecture produces descriptors invariant to $SO(2)$ rotations around the spin axis while preserving rich geometric detail, enabling robust alignment even under varying rotations. Another state-of-the-art approach, Predator by Huang \emph{et al.} \cite{Huang2021}, incorporates an attention-based mechanism to handle the case of partially overlapping point clouds. It introduces an overlap attention module that helps the network focus on points likely to have correspondences in the other cloud.

Other notable deep learning contributions include CGF (Compact Geometric Features) \cite{Khoury2017}, which used a random forest to pre-align patches and then a simple CNN, producing a 32-dimensional descriptor that was among the first learning-based methods to rival handcrafted features on standard datasets. 3DFeat-Net \cite{Yew2018} was a pioneering attempt to learn both detector and descriptor in a weakly-supervised manner (using GPS/INS for relative pose instead of exact correspondences). It targeted outdoor LiDAR point clouds and introduced an alignment loss to encourage network-predicted keypoints to match between frames. Similarly, USIP \cite{Li2019} proposed an unsupervised way to learn 3D keypoint detection by maximizing repeatability. While these methods did not always outperform handcrafted detectors in all cases, they represent important steps toward fully learning-based feature matching pipelines.

In summary, deep learning-based approaches have significantly advanced the state-of-the-art in depth image feature matching. They achieve higher matching accuracy and robustness by learning from large data corpora of 3D scenes, capturing complex geometric patterns that handcrafted descriptors struggle to encode. However, learned descriptors typically require careful training on data similar to the target domain and may generalize poorly if the sensor or environment differs (a challenge that recent works explicitly address through rotation-equivariant designs or training on diverse datasets). The trend in current research is toward end-to-end frameworks that jointly optimize keypoint detection, local description, and even matching or pose estimation. This integration is moving the field beyond treating detection/description as independent problems, towards holistic depth-based correspondence networks that maximize overall registration or recognition metrics.

\begin{table*}[t]
\small
\centering
\caption{The notable works on 3D data feature matching.}
\label{tab:top10-3d}

\begin{tabular}{p{2.2cm} p{5.4cm} p{5.5cm}}
\toprule

\textbf{Method} & \textbf{Brief Description} & \textbf{Key Advantage}\\ \midrule

Spin Images \cite{Johnson1999} &
Introduced a 2-D histogram descriptor for 3-D surfaces (spin images). &
Rotation-invariant; robust to clutter; enabled early 3-D object recognition from depth-only data. \\

3-D Shape Context \cite{Frome2004} &
Extended shape context to 3-D point clouds for range-image matching. &
Captures global point distribution in a local patch; effective on early range datasets. \\

FPFH \cite{Rusu2009} &
Fast Point Feature Histograms for surface registration. &
Efficient computation (lower complexity than PFH); robust on noisy scans; implemented in the PCL library. \\

SHOT \cite{Tombari2010b} &
Signature of Histograms of Orientations descriptor with a local reference frame. &
Highly distinctive yet fast, rotation-invariant through an accurate LRF; widely used baseline. \\

2.5-D SIFT \cite{Lo2009} &
Adapted SIFT keypoints and descriptors to depth (range) images. &
Reuses mature 2-D SIFT pipeline; invariant to in-plane transformations on depth maps. \\

3DMatch \cite{Zeng2017} &
First deep-learned local descriptor for RGB-D scan alignment. &
Data-driven; markedly improves match recall over hand-crafted features. \\

CGF \cite{Khoury2017} &
Learned Compact Geometric Features (32-D) on 3-D point clouds. &
Very low-dimensional; efficient matching with competitive accuracy—early learning-based success. \\

FCGF \cite{Choy2019} &
Fully Convolutional Geometric Features using 3-D sparse convolutions. &
End-to-end dense extraction; single fast forward pass; high accuracy on indoor \& outdoor data. \\

D3Feat \cite{Bai2020} &
Jointly learned 3-D keypoint detector and descriptor (dense CNN). &
Simultaneously optimises detection and description; high recall; task-specific keypoints improve registration. \\

PREDATOR \cite{Huang2021} &
Overlap-aware transformer for low-overlap point-cloud registration. &
Focuses on overlapping regions; state-of-the-art registration recall on challenging cases. \\ 

Coupled Laplacian \cite{bastico2024coupled} & Locally-aware rigid point cloud matching using graph Laplacian eigen maps with Coupled Laplacian operator. & Robustly captures fine local geometrical differences and handles eigen maps alignment without landmark dependency. \\

\bottomrule
\end{tabular}
\end{table*}

Table \ref{tab:top10-3d} summarizes some of the most significant works in depth image feature matching, while Table~\ref{tab:depth-3dmatch} summarizes methods' performance on the 3DMatch dataset~\cite{Zeng2017}. Early milestones like spin images, 3D shape contexts, and FPFH introduced foundational concepts for describing 3D local geometry. Later, the advent of learned descriptors (3DMatch, CGF) and advanced deep architectures (FCGF, D3Feat, Predator) greatly boosted matching performance and robustness, even in very challenging scenarios.

\subsection{Geometry-Based Cross-Modal Matching (2D--3D)}
Geometry-based 2D--3D cross-modal matching is critical for sensor fusion, localization, and robotics, and must bridge the large gap between dense image appearance and sparse, irregular 3D geometry. Typical scenarios include image-to-point-cloud registration for camera pose estimation, RGB--LiDAR fusion in autonomous driving, and matching 2D observations against a 3D map for long-term localization. A first family relies on projection-based correspondences, projecting 3D points onto the image plane or lifting 2D keypoints into 3D via depth estimates.

Beyond explicit projection, recent work exploits pretrained priors to reduce the modality gap: FreeReg~\cite{wang2023freereg} achieves zero-shot image-to-point-cloud registration by combining cross-modality diffusion features with monocular-depth geometric cues, while diffusion modeling on the $\mathrm{SE}(3)$ manifold~\cite{jiang2023se} casts point cloud registration as a denoising process for robust 6D pose estimation. A further line learns joint 2D--3D representations with cross-modal transformers—e.g., Barroso-Laguna \emph{et al.}~\cite{Barroso2024Matching3D} derive pose constraints from metric correspondences, MASt3R~\cite{Leroy2024MASt3R} aggregates multi-view appearance and geometry in a unified encoder, and RoboEye~\cite{Zhang2025RoboEye} adds selective 3D keypoint matching for robust robotic manipulation.

\subsection{Benchmark Datasets and Evaluation Protocols}
Research in 3d data feature matching has been accelerated by the availability of benchmark datasets and standardized evaluation protocols. We summarize some of the most widely used datasets for assessing 3D local features in Table \ref{tab:datasets-3d}. These benchmarks span indoor RGB-D scans, LiDAR point cloud data, and synthetic scenes, providing a comprehensive testbed for feature matching algorithms.

\begin{table*}[t]
\small
\centering
\caption{The benchmark datasets for depth‐image feature matching. Each dataset contains depth (range) data plus ground truth for assessing correspondence accuracy; common evaluation protocols are listed.}
\label{tab:datasets-3d}
\begin{tabular}{@{}p{2.0cm} p{2.0cm} p{4.1cm} p{4.8cm}@{}}
\toprule
\textbf{Dataset} & \textbf{Use–case} & \textbf{Description} & \textbf{Typical Evaluation Protocol} \\ \midrule
TUM RGB-D \cite{Sturm2012} &
RGB-D SLAM / odometry &
Indoor handheld RGB-D sequences (living-room, office, etc.) with ground-truth camera trajectories. &
Absolute / relative pose error (ATE, RPE); key-point repeatability across frames; relocalization success rate. \\

ICL-NUIM \cite{Handa2014} &
Synthetic indoor SLAM &
Rendered RGB-D images of synthetic apartments with exact pose and map ground truth. &
Same trajectory-error metrics as TUM; enables controlled, noise-free evaluation. \\

7-Scenes \cite{Shotton2013} &
Camera relocalization &
Seven small indoor scenes were recorded with RGB-D; each frame has a known pose. &
Percentage of queries within 5 cm / 5 degrees of ground truth; precision/recall of feature matches. \\

Redwood \cite{Choi2015} &
3-D local patch matching &
Pairs of point-cloud fragments from indoor scenes with known relative transforms. &
Registration recall (fraction correctly aligned using RANSAC on matches); match-recall at fixed precision. \\

ETH Laser Reg. \cite{Pomerleau2012} &
General 3-D registration &
Laser-scan datasets (office, apartment, stairwell, gazebo, forest) with ground-truth poses. &
RMS alignment error, registration success rate (converging below threshold); inlier ratio after ICP / feature matching. \\

KITTI Odometry \cite{Geiger2013} &
Outdoor LiDAR odometry &
HDL-64 LiDAR scans of driving scenes, plus GPS/INS ground-truth poses; stereo depth for some sequences. &
Trajectory drift per sequence; outlier ratio of feature matches, pairwise registration recall. \\

3DLoMatch \cite{Huang2021} &
Low-overlap registration &
Subset of 3DMatch with fragment pairs having $<30\%$ overlap. &
Registration recall under low-overlap; inlier precision/recall; runtime for partial-overlap matching. \\ \bottomrule
\end{tabular}
\end{table*}

\begin{table*}[t]
\centering
\small
\caption{3D feature matching performance on the 3DMatch benchmark~\cite{Zeng2017}.
We report Feature-Match Recall (FMR, \%, higher is better) under the standard protocol
with $f{=}5000$ sampled points, $\tau_1{=}10$cm, and $\tau_2{=}5\%$ (as in~\cite{Bai2020, Ao2021}). Unless otherwise specified, all other results are taken from publicly available benchmark papers.}
\setlength{\tabcolsep}{5pt}
\renewcommand{\arraystretch}{1.08}
\resizebox{\linewidth}{!}{%
\begin{tabular}{l l c c l c}
\toprule
Method & Category & Venue & Year & Backbone / Representation & FMR (\%) \\
\midrule
FPFH~\cite{Rusu2009}
& Handcrafted 3D descriptor
& ICRA
& 2009
& --
& 35.9 \\

SHOT~\cite{Tombari2010b}
& Handcrafted 3D descriptor
& ECCV
& 2010
& --
& 23.8 \\

3DMatch~\cite{Zeng2017}
& Learned local 3D descriptor
& CVPR
& 2017
& 3D CNN (TDF voxel patch)
& 59.6 \\

PerfectMatch~\cite{Gojcic2018}
& Learned rotation-invariant local descriptor
& CVPR
& 2019
& 3D CNN on SDV voxels (LRF-aligned)
& 94.7 \\

FCGF~\cite{Choy2019}
& SparseConv learned descriptor
& ICCV
& 2019
& SparseConv (FCGF)
& 95.2 \\

D3Feat~\cite{Bai2020}
& Joint detect+describe (learned)
& CVPR
& 2020
& Fully-conv detect+describe
& 95.8 \\

PREDATOR~\cite{Huang2021}
& Overlap-attention + keypoint selection
& CVPR
& 2021
& Overlap attention (learned keypoints)
& 96.6 \\

SpinNet~\cite{Ao2021}
& Rotation-invariant learned descriptor
& CVPR
& 2021
& Cylindrical conv on LRF-aligned patches
& 97.6 \\

CoFiNet~\cite{Yu2021_3d}
& Coarse-to-fine correspondence network
& NeurIPS
& 2021
& KPConv encoder + coarse-to-fine matching
& 98.1 \\


HST~\cite{huang2024consistency}
& Spot-guided hierarchical transformer registration
& NeurIPS
& 2024
& Transformer (hierarchical)
& 98.8 \\

Cross-PCR~\cite{zhao2025cross}
& Density-robust + loose-to-strict matching
& AAAI
& 2025
& Density-robust encoder
& 97.9 \\

\bottomrule
\end{tabular}
}
\label{tab:depth-3dmatch}
\end{table*}

%% file: 4-5-combined.tex
\section{Medical Images}

Medical image registration involves aligning multiple images into a unified spatial coordinate system, crucial for clinical and research applications such as longitudinal analysis, multi-atlas segmentation, and motion correction. This alignment spans both single-modality (e.g., MRI-to-MRI) and cross-modality (e.g., MRI-to-CT, PET-to-MRI) scenarios, each presenting unique challenges due to differences in imaging characteristics and intensity distributions.

\subsection{Handcrafted Methods}
Traditional medical registration methods primarily fall into intensity-based and feature-based approaches. Intensity-based methods directly optimize voxel-wise similarity measures, initially employing simple metrics like sum-of-squared differences or cross-correlation for same-modality alignment \cite{brown1992survey,Hill2001}. A significant advancement was Mutual Information (MI), introduced independently by Maes \textit{et al.} and Viola \& Wells \cite{viola1997alignment,Maes1997}, becoming the standard for cross-modality registration due to its robustness against varying intensity distributions between modalities. Variants like Normalized Mutual Information (NMI) further enhanced performance \cite{studholme1999nmi}.

Feature-based methods rely on identifying and matching salient anatomical landmarks, edges, or surfaces. Landmark-based registration methods, such as thin-plate splines (TPS), precisely align identified points \cite{bookstein1989tps}, whereas surface-based methods like the Iterative Closest Point (ICP) algorithm are effective for rigid alignments \cite{besl1992icp}. Modality-independent descriptors, notably the Modality Independent Neighbourhood Descriptor (MIND) \cite{Heinrich2012}, significantly advanced cross-modality feature matching by encoding local image structure robustly across diverse modalities.

To ensure physically plausible deformations, diffeomorphic methods like Large Deformation Diffeomorphic Metric Mapping (LDDMM) \cite{beg2005lddmm} and Symmetric Normalization (SyN) \cite{avants2008symmetric} were developed. Widely adopted software toolkits, including Elastix \cite{klein2010elastix} and ITK \cite{yushkevich2016itk}, encapsulate these robust registration methodologies, facilitating broader clinical and research adoption.

\subsection{Deep Learning-Based Methods}
The emergence of deep learning has profoundly impacted medical image registration by automating feature extraction and transformation prediction. Initial supervised deep methods required labeled deformation fields \cite{sokooti2017,yang2017quicksilver}, but unsupervised learning paradigms, notably exemplified by VoxelMorph \cite{balakrishnan2019voxelmorph}, allowed for training without explicit ground truth, optimizing similarity metrics directly via differentiable spatial transformations.

Later, deep architectures evolved, introducing multi-stage cascade networks for handling large deformations effectively \cite{zhao2019VTN,eppenhof2019,jiang2022one}. Crucially, enforcing diffeomorphism within deep networks, using frameworks like Laplacian Pyramid Network (LapIRN) \cite{mok2020lapirn} and VoxelMorph-diffeomorphic \cite{dalca2019}, integrated theoretical robustness with computational efficiency. Neural ordinary differential equations (ODEs) and implicit representations further advanced these frameworks by ensuring smooth, invertible transformations \cite{xu2021neurODE,han2023dnvf}.

Adversarial training provided another avenue for enhancing registration accuracy by leveraging discriminators to critique alignments and implicitly learn robust similarity metrics \cite{fan2018adversarial,mahapatra2020ganreg,elmahdy2019,wang2024superjunction,zeng2025towards}. Transformers introduced powerful mechanisms to capture long-range spatial dependencies effectively, as illustrated by ViT-V-Net \cite{chen2021vitvnet}, TransMorph \cite{chen2022transmorph}, and XMorpher \cite{shi2022xmorpher}, demonstrating improved global registration performance.

Cross-modality deep learning approaches often utilize modality translation as an intermediate step via generative adversarial networks (GANs) and cycle-consistency frameworks, transforming the multi-modal registration problem into a mono-modal scenario \cite{Cao2017,yang2022cross,Mahapatra2018}. Moreover, modality-invariant embeddings and cross-modal attention mechanisms directly facilitate accurate multi-modal alignment without explicit modality conversion \cite{Simonovsky2016,Song2022,Wang2022crossmodality}.

Emerging trends include employing diffusion models, such as DiffuseMorph \cite{kim2022diffusemorph}, to inherently model deformation uncertainties, and exploring implicit neural representations for continuous spatial transformations \cite{han2023dnvf}. These methods highlight potential advantages in uncertainty modeling and diverse deformation sampling, though further investigation is needed.

The integration of classical rigor and deep learning's flexibility is reshaping medical image registration. While handcrafted methods remain valuable for their interpretability and theoretical guarantees, deep learning methods offer unprecedented speed and adaptability to complex scenarios. 
Table~\ref{tab:top10papers} summarizes some landmark methods in the development of medical image registration, spanning from the introduction of mutual information to recent deep learning innovations. Each has significantly influenced this field. Table~\ref{tab:medical-learn2reg} summarizes methods performance on the Learn2Reg dataset~\cite{Hering2022}.

\begin{table*}[t!]
\caption{The influential methods on medical image registration. Each paper introduced a key concept or method; advantages are noted in context. \label{tab:top10papers}}
\renewcommand{\arraystretch}{1.15}
\footnotesize
\centering
\begin{tabular}{p{1.6cm} p{5.7cm} p{5.7cm}}
\hline
\textbf{Method} & \textbf{Brief Description} & \textbf{Key Advantage} \\
\hline 
MI \cite{viola1997alignment} & Introduced mutual information (MI) for multi-modal alignment; maximization of MI to find the best transform. Demonstrated registration of CT/MRI with no prior segmentation. & Landmark method enabling fully automatic multi-modal registration; invariant to intensity scaling; became the foundation for most intensity-based multi-modal registrations. \\

NMI \cite{studholme1999nmi} & Proposed Normalized Mutual Information (NMI) to improve robustness of MI by compensating for overlap changes. Applied to 3D brain registration. & NMI handled cases with limited overlap and was less biased by areas outside common volume; it became standard in applications like brain MRI-PET alignment. \\

Maes \textit{et al.} \cite{Maes1997} & Simultaneous, independent introduction of MI-based registration (with normalized MI variants) and validation on brain CT/MR and PET/MR datasets. & Reinforced MI as a modality-agnostic similarity. Their software became widely adopted; it showed sub-voxel accuracy comparable to experts. \\

SyN \cite{avants2008symmetric} & Developed SyN (Symmetric Normalization) diffeomorphic registration. Used cross-correlation as similarity (robust for multi-modal MRI). Showed excellent results on inter-subject brain MRI (evaluated on cross-modality too). & SyN provided very accurate deformable registration with theoretical invertibility. The approach (implemented in ANTs) became top-ranked in many benchmarks and is applicable to multi-modal (with CC or MI). \\
MIND \cite{Heinrich2012} & Introduced MIND descriptor for multi-modal deformable registration. Demonstrated MIND-based alignment on CT/MR chest and brain, outperforming MI on challenging cases. & Provided a lightweight, modality-independent feature that could plug into any registration algorithm; improved robustness to intensity distortions and outliers (e.g., pathology). \\
Simonovsky \textit{et al.} \cite{Simonovsky2016} & Learned a CNN-based metric for multi-modal (MR-CT) patch similarity. Integrated this learned metric into a registration framework (replacing MI). & Pioneered the use of deep learning to directly improve the similarity measure. Achieved higher accuracy than handcrafted metrics, opening the door to learning-based similarity in registration. \\
VoxelMorph \cite{balakrishnan2019voxelmorph} & VoxelMorph unsupervised learning framework for deformable registration. Trained on MRI (later extended to CT, etc.) using spatial transformer networks and a reconstruction loss. & Extremely fast at test time (CNN predicts deformation in one pass); demonstrated learning-based registration can approach the quality of iterative methods while being orders of magnitude faster. \\
Fan \textit{et al.} \cite{Fan2019} & Adversarial learning for multi-modal (and mono-modal) registration. Used a generator network for deformable registration and a discriminator to judge alignment realism between MR and CT. & One of the first to apply GANs to registration. Improved alignment in cross-modality by learning implicit common representations; showed potential of adversarial loss to capture complex appearance differences. \\
\hline
\end{tabular}
\end{table*}

\subsection{Benchmark Datasets and Evaluation Protocols} 

Research in medical image registration relies on public datasets and standardized protocols, which are especially important for multi-modal methods where ground-truth ``known'' transformations are rarely available. Table~\ref{tab:top10datasets} summarizes the most widely used datasets, their modalities, use cases, and evaluation protocols. Foundational resources include RIRE~\cite{West1997}, which uses fiducial markers on multi-modal brain scans (CT/MR/PET) to provide gold-standard transforms for target registration error (TRE) evaluation; IXI, with multi-sequence MRI (T1/T2/PD) of healthy subjects for cross-modality experiments; and BraTS~\cite{Menze2015}, whose multi-parametric tumor MRI is used to assess deformable inter-sequence alignment.

When ground-truth alignment is unavailable, a common protocol uses surrogate ground truth from anatomical labels: the candidate registration propagates a segmentation across images, and overlap (Dice) with a manual segmentation is measured, as in the MM-WHS challenge~\cite{Zhuang2019} for cardiac MRI--CT label transfer. The learning era brought community challenges such as Learn2Reg~\cite{Hering2022}, spanning cross-modality tasks (CT--MRI abdomen, MR--ultrasound brain) and establishing modern evaluation protocols: TRE on landmarks, Dice on propagated labels, inverse-consistency error, Jacobian-determinant analysis of non-physical folds, and run-time/memory comparisons.

\begin{table*}[t!]
\small
\caption{The public datasets for medical image registration, with modality details, use cases, and evaluation notes.
\label{tab:top10datasets}}
\centering
\renewcommand{\arraystretch}{1.1}

\begin{tabular}{@{}p{1.5cm} p{1.5cm} p{1.5cm} p{8.5cm}@{}}
\hline
\textbf{Dataset} & \textbf{Modality} & \textbf{Use-Case} & \textbf{Description \& Evaluation Protocol} \\
\hline 
RIRE \cite{West1997} & CT, MR, PET (Brain) & Rigid reg., algorithm comparison & Retrospective Image Registration Evaluation: Brain images with implanted markers. Gold-standard rigid transforms known via fiducials. Error measured as TRE (mm) at marked anatomical targets; enabled first objective ranking of multi-modal reg methods. \\
IXI \cite{ixi_dataset_2025}  & MRI (T1, T2, PD) (Brain)& Deformable reg., synthesis & Large collection of normal brain MRI in multiple sequences for each subject. Used for cross-sequence registration and learning modality mappings. Evaluation by mutual information or downstream task (e.g., segmentation consistency). \\
BraTS \cite{Menze2015} & MRI (T1, T1c, T2, FLAIR) (Brain) & Deformable reg., tumor analysis & Brain Tumor Segmentation Challenge data. Multi-parametric MR per patient (already roughly aligned). Registration methods tested by aligning sequences or longitudinal scans; evaluation via overlap of tumor/structure labels or visual consistency of tumor boundaries. \\
MM-WHS \cite{Zhuang2019} & MRI, CT (Heart) & Cross-modality segmentation (indirect reg) & Multi-Modality Whole Heart Segmentation Challenge: 60 cardiac CT and 60 MRI from different patients with labeled structures. Not paired, but used to test reg-by-seg approaches and domain adaptation. Reg algorithms can map an atlas from CT to MRI (or vice versa); evaluated by Dice of propagated labels and surface distances. \\
Learn2Reg \cite{Hering2022} & CT/MR (abdomen), MR/US (brain), others & Multiple (challenge tasks) & Challenge with multiple multi-modal registration tasks: e.g., intra-patient liver CT-MRI, neurosurgical MR-US. Provided training and test splits. Evaluation via hidden ground truth or expert landmark analysis; metrics included TRE on hidden landmarks, Dice of organ masks, and fold count in deformations. Established a common benchmark for learning-based methods. \\
\hline
\end{tabular}
\end{table*}

\subsection{Applications}
Cross-modality registration underpins several clinical workflows. In \textbf{diagnostic imaging fusion}, registering brain MRI and PET combines anatomical detail with metabolic information, letting clinicians localize PET findings such as epileptic foci or tumor activity on MRI—typically via rigid MI-based registration, now routine in brain tumor and dementia evaluations \cite{Somer2003}. CT–MRI registration is likewise central to radiotherapy planning for head, neck, and pelvic cancers, where MRI delineates soft-tissue extent and CT supplies electron density; deformable multi-modal algorithms are needed around organs that deform, with alignment accuracy directly affecting treatment margins.

In \textbf{image-guided surgery}, pre-operative MRI/CT must be aligned to intra-operative imaging. A prominent case is MRI-to-ultrasound registration in neurosurgery, where deformable alignment compensates for brain shift after skull opening to update neuro-navigation; MIND descriptors or learned features are commonly used, supported by databases such as RESECT and BITE. Similarly, MR–TRUS registration maps MRI-identified lesions onto real-time ultrasound for targeted prostate biopsy. A third use is \textbf{multi-modal atlas-based segmentation}: registering delineated atlases across modalities propagates organ or pathology labels—e.g., Ding \textit{et al.}~\cite{Ding2022} warped CT atlases to MR for cardiac labeling with near same-modality accuracy. Overall, cross-modality registration is now embedded in neuro-navigation, radiotherapy, and augmented-reality surgical guidance.

\begin{table*}[t]
\centering
\small
\caption{Brain MRI inter-patient registration on Learn2Reg. Performance is measured by mean Dice similarity (\%, higher is better). Results are taken from public benchmark reports / challenge papers.}
\setlength{\tabcolsep}{4pt}
\renewcommand{\arraystretch}{1.08}
\resizebox{\linewidth}{!}{%
\begin{tabular}{l l c c c  c}
\toprule
\textbf{Method} & \textbf{Category} & \textbf{Venue} & \textbf{Year} & \textbf{Backbone} & \textbf{Dice (\%)} \\
\midrule
SynthMorph~\cite{hoffmann2021synthmorph}
& CNN registration trained via synthetic data (contrast-invariant)
& TMI
& 2021
& U-Net-style CNN 
& 72.43 \\

VoxelMorph~\cite{balakrishnan2019voxelmorph}
& CNN-based unsupervised deformable registration
& TMI
& 2019
& U-Net CNN + spatial transformer
& 73.25 \\

TransMorph~\cite{chen2022transmorph}
& Transformer-based unsupervised registration
& MedIA
& 2022
& Transformer-ConvNet hybrid
& 75.94 \\

Zhang \emph{et al.}~\cite{zhang2024large}
& Challenge method (efficient CNN registration backbone)
& arXiv
& 2024
& CNN backbone + large-kernel conv 
& 77.34 \\

EOIR~\cite{chen2025encoder}
& Lightweight encoder-only registration
& arXiv
& 2025
& Encoder-only multi-scale design
& 77.37 \\
\bottomrule
\end{tabular}
}
\label{tab:medical-learn2reg}
\end{table*}

%% file: 6_cross_vision_language.tex
\section{Cross-Modality Vision-Language Feature Matching: From Traditional Methods to Large-Scale Foundation Models}
Vision-language feature matching refers to the broad set of techniques that enable joint understanding of visual content and natural language. In recent years, a variety of tasks have emerged at the intersection of computer vision and natural language processing, requiring models to align and integrate visual and textual information \cite{Radford2021}. Such tasks include image captioning \cite{showtell15}, visual question answering (VQA) \cite{vqa15}, cross-modal retrieval \cite{karpathy15}, among others. 

Progress in these areas has been propelled by advances in deep learning, the development of large-scale multimodal datasets, and the introduction of powerful vision-language pre-training paradigms. This section provides a comprehensive review of the field, covering the major areas: (1) cross-modal retrieval and search; (2) visual grounding and referring expressions; (3) open-vocabulary classification, detection, and segmentation; (4) image captioning; and (5) benchmark datasets and evaluation protocols. In addition, we identify the particularly pivotal works across all topics and summarize them in Table~\ref{tab:top10-cross-vision_language}, and we provide an overview of important datasets in Table~\ref{tab:datasets-cross-vision-language}. 

Compared to the geometric feature matching tasks discussed in previous sections (2-4), which aim to establish explicit spatial correspondences such as `this pixel in view~A matches that pixel in view~B'', in this section, the methods operate predominantly in a semantic space. Here, the critical question is whether ``this image or region expresses the same concept as this piece of text''. Representative approaches include scalable image--text retrieval adaptation (e.g., Multiway-Adapter~\cite{Long2024MultiwayAdapter}) and instruction-tuned vision--language models that follow free-form prompts (e.g., InstructBLIP~\cite{dai2023instructblip}). We nevertheless view vision--language matching as part of the broader feature-matching landscape because it relies on many of the same underlying mechanisms: learned embeddings, contrastive objectives, and cross-modal attention, while extending them to richer reasoning and open-world generalization.

\subsection{Cross-Modal Retrieval and Search}
A breakthrough in cross-modal retrieval came with contrastively trained dual-encoder models on web-scale data. CLIP used 400 million image-text pairs for contrastive learning, producing an image encoder and a text encoder whose embeddings are directly comparable via cosine similarity \cite{Radford2021}. CLIP achieved remarkable zero-shot retrieval and zero-shot classification performance, far surpassing previous supervised models when scaled. Concurrently, ALIGN trained on 1.8 billion noisy image-alt-text pairs and similarly found that simple dual-encoder architectures can be extremely effective given enough data \cite{Jia2021}. These models revolutionized cross-modal retrieval by enabling open-domain searches: e.g., using natural language queries to find images in an unseen collection. 
Recent research in cross-modal retrieval has explored hybrid approaches that combine the efficiency of dual encoders with the precision of interaction models. Examples include leveraging CLIP features with lightweight cross-attention re-ranking to refine results \cite{Lei2021}, or learning modality-aware experts and fusion at retrieval time \cite{Zhong2022}. Nonetheless, the dominant trend is using large pre-trained vision-language models and fine-tuning or prompt-tuning them for retrieval tasks. 

\subsection{Visual Grounding and Referring Expressions}
Early work on referring expressions built upon object detection and language models. Initial approaches often used a two-stage pipeline: generate region proposals, then rank them by how well they match the query expression. For instance, Mao \etal \cite{Mao2016} presented a CNN-LSTM model that could both generate and comprehend referring expressions. 

Subsequent methods improved upon visual and linguistic feature fusion. Hu~\etal \cite{Hu2016} proposed to incorporate context such as surrounding objects' features to disambiguate references. Speaker-Listener-Reinforcer models \cite{Yu2016} jointly trained a generator (speaker) and a comprehension model (listener) with reinforcement learning to ensure that generated expressions are discriminative and comprehensible. This approach improved grounding accuracy by using the listener’s feedback as a reward for the speaker, effectively pushing the descriptions (and the comprehension) to focus on uniquely identifying details. Later, MAttNet (Modular Attention Network) achieved state-of-the-art on RefCOCO by decomposing the expression into subject, location, and relationship components, each attended by a separate module that guided the visual feature processing \cite{Yu2018}. MAttNet explicitly handled attributes (e.g., color or size), relations (e.g., “next to the table”), and the target object itself, combining the evidence to score proposals. This modular design significantly improved grounding, especially for complex expressions.

One-Stage Grounding methods removed the need for an explicit proposal stage by directly predicting the box coordinates from language and image features \cite{Yang2019}. They formulated grounding as a regression problem conditioned on text, using techniques from one-stage detectors to output the referred object location in a single forward pass. At the same time, large multi-task vision-language models started to incorporate grounding capabilities. For example, the pre-trained VilBERT~\cite{Lu2019} and UNITER~\cite{Chen2020} models could be fine-tuned for referring expression tasks by adding an output layer that selects the region corresponding to the expression \cite{Chen2020,Lu2019}. These models leverage cross-modal attention to directly align words with image regions represented by object detection features (such as Faster R-CNN region features).
In the same vein, LXMERT~\cite{tan2019lxmert} is an early and influential cross-modality pre-training framework that explicitly separates language encoding, object-relation encoding, and a cross-modality encoder to model fine-grained vision-language alignment. Its multi-task pre-training established a strong recipe for transferring transformer-based fusion models to grounding- and reasoning-heavy tasks.

The introduction of transformer-based detection models further revolutionized visual grounding. 
Beyond region-feature-based fusion, ViLT~\cite{kim2021vilt} demonstrated that competitive vision-language understanding can be achieved with a minimal transformer that directly tokenizes image patches and text tokens, removing the need for convolutional backbones and region-level supervision. 
MDETR is a notable example that unified object detection and grounding in one end-to-end transformer model \cite{Kamath2021}. MDETR extended the DETR object detector by feeding the text encoding into the transformer decoder alongside visual features. 
Following MDETR, other works have improved grounded detection. Such as TransVG also used a pure transformer encoder-decoder for visual grounding, with careful feature fusion between a vision transformer and language embeddings \cite{deng2021transvg}. Referring Transformer likewise applied a transformer architecture specialized for grounding tasks \cite{LiSigal2021}. These models benefit from the global context and multi-head attention of transformers to resolve ambiguous language by looking at all objects simultaneously and attending to relevant parts of the image for each phrase in the query.

Another line of advancement is large-scale grounded pre-training. GLIP (Grounded Language-Image Pre-training) treated every object detection training example as a phrase grounding example by converting class labels to words, and trained a model to align region proposals with those words \cite{Li2022}. By doing so on millions of examples, GLIP learned a unified representation for detection and grounding.

\subsection{Open-Vocabulary Recognition, Detection, and Segmentation}
A significant challenge in computer vision is recognizing and localizing visual concepts that were not seen during training. Traditionally, vision models were limited to closed vocabularies (fixed sets of classes or labels). However, by leveraging semantic information from language, models can extend their knowledge to an open vocabulary.

\subsubsection{Open-Vocabulary Image Recognition}
Early work in zero-shot image classification predated the deep learning wave and often relied on human-defined attributes or semantic word embeddings. Lampert et al.  \cite{Lampert2009} introduced an attribute-based classification approach: models were trained to predict intermediate attributes (like “has stripes” or “has fur”) on seen classes and then infer unseen classes by their attribute signature. This idea of between-class attribute transfer allowed recognition of new categories (e.g., “zebra”) by reasoning about attributes (striped, four-legged, etc.) even if no zebra images were in the training set. Later, Socher et al. proposed mapping image features and class name embeddings (obtained from text corpora) into a common space, enabling zero-shot recognition by selecting the class whose embedding is closest to the image embedding \cite{Socher2013}. Norouzi et al. extended this by introducing ConSE, which averaged the embeddings of predicted seen classes to synthesize an embedding for the image and compared it to unseen class embeddings \cite{Norouzi2014}.

Following CLIP, various improvements and adaptations emerged: ALIGN \cite{Jia2021} similarly learned dual encoders on an even larger dataset; LiT explored locking the text encoder while fine-tuning the image encoder for better transfer \cite{Zhai2022}; FILIP introduced finer-grained alignment at the token level, attempting to match words to image patches for improved zero-shot recognition \cite{Yao2021}. Another extension is prompt engineering and prompt learning: instead of using a simple prompt like “a photo of a [class],” methods like CoOp learn continuous prompt vectors to condition CLIP for a given downstream classification task, yielding better performance, especially in the few-shot regime \cite{Zhou2022-CoOp}. These approaches demonstrate the flexibility of open-vocabulary classifiers: since the image encoder is fixed, one can adapt the textual side to different tasks or domains with minimal effort.

\subsubsection{Open-Vocabulary Object Detection}
Extending open-vocabulary recognition to object detection is more complex, as it requires localizing unseen classes in images. Early zero-shot detection approaches adapted zero-shot classification techniques to the detection pipeline. For instance, Bansal et al. proposed one of the first zero-shot object detection methods by incorporating class semantic embeddings from Word2Vec~\cite{mikolov2013efficient} or GloVe \cite{pennington2014glove} into the network and formulating a loss that encourages detection boxes to predict those embeddings for unseen classes \cite{Bansal2018}. They often relied on attribute predictions or careful handling of background regions to avoid confusion. 

However, like classification, open-vocabulary detection saw major progress with the advent of powerful vision-language models. One line of work leveraged the region classification head of a detector. For example, Kim et al. introduced Region-aware Open-vocabulary ViT, which is trained on paired image-caption data to detect objects described in captions, thereby learning to localize a wider variety of concepts than the fixed label set of detection datasets \cite{kim2023region}. 

Another important work is ViLD \cite{gu2021open}, which took a standard detector (like Faster R-CNN) and replaced or augmented its classifier with CLIP’s image-text similarity: the region-of-interest features from the detector were matched to text embeddings of class names. Essentially, they transferred the classifier’s knowledge to an open vocabulary by using CLIP’s text encoder as the classification layer.
Similarly, RegionCLIP went a step further by also fine-tuning the image encoder at the region level: it generated region proposals and paired them with caption segments during training to better align region features with text \cite{Zhong2022}. By iteratively refining region-text alignment, RegionCLIP improved open-vocabulary detection performance, especially on smaller or harder-to-recognize objects.

Another family of approaches trains detectors on the fly with language. For example, GLIP \cite{Li2022}, as mentioned, unified detection and grounding: it can take arbitrary text queries and highlight those objects in the image. GLIP’s training formulated detection as phrase grounding; thus it naturally handles open vocabulary by taking the object’s category name as a phrase. OWL-ViT \cite{Minderer2022} similarly built on a Vision Transformer to detect objects described by text prompts, enabling flexible queries like “a decorated cake” and returning matching boxes. 
More recently, Grounding DINO~\cite{liu2024grounding} advanced open-set detection by combining a strong transformer detector (DINO-style detection) with grounded pre-training, enabling detection of arbitrary categories or referring expressions specified by free-form text.
Complementarily, Scaling Open-Vocabulary Object Detection~\cite{minderer2023scaling} systematically studied how to push open-vocabulary detection toward web-scale training via self-training on image-text data. The work introduces the OWLv2 model and an OWL-ST recipe that generates pseudo-box annotations at scale and filters them efficiently.

\subsubsection{Open-Vocabulary Semantic Segmentation}
Earlier zero-shot segmentation methods often extended zero-shot classification by using word embeddings of class names and relating them to pixel-level features. Bucher \textit{et al.} \cite{Bucher2019} proposed a zero-shot segmentation approach that projected image features and class embedding vectors into a common space and computed segmentation masks for unseen classes via similarity. 
Recently, open-vocabulary segmentation has benefited from vision-language models and multi-modal training. One notable approach is LSeg \cite{li2022language}. LSeg introduced a transformer-based model that takes an arbitrary text label and produces segmentation masks. 
It uses a contrastive training objective to align pixel embeddings with text embeddings for known classes, which encourages pixels of unseen classes to naturally align with semantically similar text descriptions. 
Another approach is GroupViT, which combined vision transformers with CLIP pretraining to perform segmentation via a grouping process \cite{xu2022groupvit}. 
OpenSeg explored training a segmentation model on the union of many segmentation datasets plus image-caption data, using a text encoder to represent class labels and even long descriptions for each mask \cite{ghiasi2022scaling}. Similarly, the trend of using CLIP's semantic space has influenced segmentation: for example, CLIPSeg fine-tuned CLIP to produce segmentation masks given a text prompt, essentially performing text-conditioned segmentation in a zero-shot manner \cite{Luddecke2022}.

In conclusion, open-vocabulary recognition (classification, detection, segmentation)~\cite{liu2025physically} is a clear beneficiary of vision-language feature alignment. By training on diverse data or explicitly coupling vision models with language embeddings, we can achieve recognition of a virtually unlimited set of concepts. Table~\ref{tab:datasets-cross-vision-language} lists key datasets, typically with modified evaluation protocols to test generalization to unseen classes.

\subsection{Image Captioning}
Image captioning aims to generate a natural language description of an image. Show and Tell \cite{Vinyals2015} is a seminal CNN--LSTM model that encodes an image as a feature vector and decodes captions word by word. Show, Attend and Tell introduced spatial attention into caption generation, learning to focus on different image regions at each decoding step \cite{Xu2015}. \\
Subsequent advances improved both visual features and training objectives. Stronger backbones (e.g., ResNet/EfficientNet) and region-based features from \cite{Anderson2018-bottom} enhanced visual representations. Self-Critical Sequence Training (SCST) directly optimized the CIDEr metric via reinforcement learning, yielding more diverse and descriptive captions \cite{Rennie2017}. 
Pre-trained transformer models then incorporated explicit object tags. OSCAR added detected object tags as input tokens during pre-training and fine-tuning \cite{Li2020Oscar}, and its successor VinVL further improved visual features, achieving state-of-the-art MSCOCO results \cite{Zhang2021}.

Recent work unifies captioning with broader vision--language tasks using large generative models. SimVLM treats captioning as language modeling with an image prefix, pre-trained end-to-end on large-scale image--text data \cite{Wang2021SimVLM}. PaLI scales this approach to billions of parameters and multilingual data \cite{chen2022pali}. BLIP-2~\cite{Li2023BLIP2} connects vision encoders with large language models, enabling flexible prompting for fluent, contextually rich captions or answers that combine visual content with world knowledge.
Instruction-tuned multimodal language models, such as InstructBLIP~\cite{dai2023instructblip}, further unify a wide range of vision--language tasks (captioning, VQA, retrieval, and grounded reasoning) under a common generative interface, improving generalization to novel instructions and domains. Embodied models like PaLM-E~\cite{driess2023palm} extend this paradigm to robotics and navigation, treating visual observations, language commands, and state information as input tokens to a single large model.

\begin{table*}[!htbp]\small
\centering
\small
\caption{The vision-language datasets with their typical use-cases, content description, and evaluation protocols. These benchmarks have driven the development of methods discussed in this survey.}
\label{tab:datasets-cross-vision-language}

\begin{tabular}{p{1.1cm} p{1.3cm} p{4.3cm} p{5.8cm}}
\toprule
\textbf{Dataset} & \textbf{Use-case} & \textbf{Description} & \textbf{Evaluation Protocol} \\
\midrule
MSCOCO Captions \cite{Chen2015} & Image Captioning, Retrieval & 123,000 images (COCO) each with 5 human-written captions. Diverse everyday scenes with people, objects, and activities. & Caption: Automatic metrics (BLEU, METEOR, ROUGE, \textbf{CIDEr}); human evaluation for fluency/accuracy. Retrieval: Recall@1,5,10 on 1K test images (e.g., COCO Karpathy split). \\
Flickr30k \cite{Young2014} & Captioning, Image-Text Retrieval & 31,000 web images with 5 captions each. Often single-object or simple scenes (people and animals in various activities). & Retrieval: Train/test sets with standard 1K test images, measure Recall@K. Captions: similar automatic metrics as COCO (though COCO is primarily for captioning). \\
RefCOCO \cite{Yu2016,Mao2016} & Referring Expression Comprehension & Images from COCO with 50k+ referring expressions for 20k objects. RefCOCO/RefCOCO+ focus on multiple objects, short phrases; RefCOCOg has longer, more complex expressions. & Accuracy of selecting the correct region given an expression. Evaluated on val/test splits, sometimes split by whether multiple instances of the object class are present. \\
VQA \cite{goyal2017making} & Visual Q\&A & $\sim$204,000 images (COCO) with 1.1M questions, each with 10 free-response answers (to reduce guess bias). Questions across categories: “What is…?”, “How many…?”, etc. & Accuracy computed by agreement with human answers: an answer is correct if at least 3 of 10 annotators gave that answer. Report overall accuracy and per-question-type accuracy on test set. \\

LVIS \cite{gupta2019lvis} & Large-Vocabulary Detection, Segmentation & 164k images (subset of COCO) with 1203 object categories (long-tail distribution). Each image has segmentation masks for present objects. & Detection/Seg: Average Precision (AP) computed for all classes, as well as AP on frequent, common, and rare splits of classes. Open-vocabulary detection methods often train on base classes and evaluate on rare (unseen) classes of LVIS to measure generalization. \\
\bottomrule
\end{tabular}
\end{table*}

\begin{table*}[!htbp]\small
\centering

\caption{The vision-language methods across different topics, with a brief description.}
\label{tab:top10-cross-vision_language}
\begin{tabular}{p{1.8cm} p{5.7cm} p{4.8cm}}
\toprule
\textbf{Paper} & \textbf{Description} & \textbf{Key Advantage} \\
\midrule
Karpathy et al. \cite{karpathy15} & Introduced a deep neural model aligning image regions and words for caption generation and retrieval. & Pioneered joint vision-language embeddings; enabled bidirectional image-caption retrieval on Flickr/MSCOCO. \\
Show, Attend and Tell \cite{Xu2015} & Applied attention mechanisms to image captioning (CNN + RNN) to focus on relevant image parts. & First use of visual attention in captioning; improved descriptive detail and interpretability of image descriptions. \\
VQA v1 \cite{vqa15} & Created the first large Visual Question Answering dataset and baseline model combining CNN and LSTM features. & Established VQA as a benchmark task; spurred extensive research into vision-language reasoning. \\
Bottom-Up and Top-Down Attention \cite{Anderson2018-bottom} & Used object detection features and two-stage attention for captioning/VQA. & Significantly improved VQA and captioning performance by leveraging detected objects as visual tokens. \\
Room-to-Room (R2R) \cite{Anderson2018-nav} &  Introduced vision-language navigation with the agent following natural-language instructions in real scenes. & Launched embodied vision-language navigation research; provided a benchmark for grounded instruction following. \\
ViLBERT \cite{Lu2019} & Proposed a two-stream transformer pre-trained on image-text data for downstream VQA, captioning, etc. & One of the first vision-language BERT models; demonstrated the power of large-scale multimodal pre-training. \\
UNITER \cite{Chen2020} & Unified single-stream transformer for vision-language tasks with extensive pre-training on image-text pairs. & Achieved state-of-the-art on multiple tasks (VQA, retrieval, grounding) by joint encoding of images and text in one transformer. \\
CLIP \cite{Radford2021} & Learned visual and textual encoders jointly via contrastive learning on 400M image-text pairs (web data). & Produced a powerful generic representation; enabled open-vocabulary image recognition and retrieval with zero-shot transfer. \\
MDETR \cite{Kamath2021} & Proposed a transformer that modulates DETR with text to directly ground expressions in images (object detection + language). & Unified referring expression comprehension with object detection; achieved strong phrase grounding by end-to-end training on aligned data. \\
GLIP \cite{Li2022} & Unified object detection and phrase grounding in a pre-training framework on aligned visual-language data. & Enabled open-vocabulary detection; excelled in detecting and localizing novel objects described by text without task-specific training. \\

DeepSeek-VL \cite{lu2024deepseek} & Open-source vision-language model featuring a hybrid vision encoder, modality-balanced pretraining, and real-world-aligned supervised fine-tuning. & Demonstrates robust multimodal understanding and superior performance in real-world scenarios and across diverse vision-language benchmarks. \\
\bottomrule
\end{tabular}
\end{table*}

\subsection{Benchmark Datasets and Evaluation Protocols}
Benchmarks have played a crucial role in driving progress in vision-language research, while Table~\ref{tab:visionlang-coco} summarizes method performance on the MSCOCO dataset~\cite{mscoco14} for the Image-text retrieval task. Table~\ref{tab:datasets-cross-vision-language} summarizes some of the most important datasets across the tasks discussed, along with their typical use-cases and evaluation metrics.

\begin{table*}[h]
\centering
\small
\caption{Image-text retrieval results on MSCOCO (5K test set)~\cite{mscoco14}. We report Recall@1 (R@1) for image-to-text (I2T) and text-to-image (T2I). Unless noted, results are COCO fine-tuned on the 5K test protocol as reported in the original papers; CLIP is reported in the zero-shot setting.}
\setlength{\tabcolsep}{5pt}
\renewcommand{\arraystretch}{1.08}
\resizebox{\linewidth}{!}{%
\begin{tabular}{l|lccc c c}
\toprule
\textbf{Method} &
\textbf{Category} &
\textbf{Venue} &
\textbf{Year} &
\textbf{Params} &
\textbf{I2T R@1} &
\textbf{T2I R@1} \\
\midrule
VSE++~\cite{faghri2017vse++} &
Dual-encoder embedding; hard-negative ranking loss (ResNet, FT; 5K test) &
BMVC & 2018 & -- & 41.3 & 30.3 \\
SCAN~\cite{Lee2018} &
Interaction model; stacked cross-attention matching (5K test) &
ECCV & 2018 & -- & 50.4 & 38.6 \\
VSRN~\cite{Li2019-img-text} &
Interaction + reasoning; visual semantic relation reasoning (5K test) &
ICCV & 2019 & -- & 53.0 & 40.5 \\
UNITER (Large)~\cite{Chen2020} &
Pretrained VL transformer; joint cross-modal encoding (fusion encoder) &
ECCV & 2020 & 303M & 65.7 & 52.9 \\
ALIGN~\cite{Jia2021} &
Dual-encoder contrastive pretraining; COCO fine-tuned retrieval &
ICML & 2021 & 820M & 77.0 & 59.9 \\
CLIP~\cite{Radford2021} &
Dual-encoder contrastive pretraining; \emph{zero-shot} retrieval on COCO-5K &
ICML & 2021 & 428M & 58.4 & 37.8 \\
BLIP-2 (ViT-g)~\cite{Li2023BLIP2} &
Foundation VLP; Q-Former bridging frozen vision encoder and LLM &
ICML & 2023 & 1.2B & 85.4 & 68.3 \\

ALBEF~\cite{li2021align} &
Align-before-fuse; momentum distillation; hybrid align+fusion training &
NeurIPS & 2021 & 210M & 77.6 & 60.7 \\

BLIPViT-L~\cite{li2022blip} &
Larger vision backbone (ViT-L) for retrieval fine-tuning &
ICML & 2022 & 470M & 82.4 & 65.1 \\

MaxMatch~\cite{alomari2025maximal} &
Set-based multi-embedding retrieval; maximal pair assignment similarity + diversity losses &
ACL & 2025 & -- & 59.5 & 42.3 \\

D2S-VSE~\cite{liu2025aligning} &
VSE with dense-to-sparse feature distillation to enrich sparse-text semantics &
ICCV & 2025 & -- & 60.1 & 46.3 \\

\bottomrule
\end{tabular}
}
\label{tab:visionlang-coco}
\end{table*}

%% file: 7_discussion_conclusion.tex
\section{Cross-cutting Challenges and Modality-Dependent Assumptions}
Across RGB, 3D, medical, and vision--language matching, several challenges are shared (e.g., outliers/occlusions, long-range ambiguities, domain shifts, and scalable evaluation), but each modality imposes distinct assumptions that strongly shape method design.
For instance, RGB matching often relies on local photometric cues and approximate viewpoint invariances; 3D matching relies on geometric consistency and rigidity priors; medical registration must handle intensity non-correspondence and non-rigid anatomical deformation; and vision--language matching emphasizes semantic compositionality, grounding, and open-vocabulary generalization.

\section{Conclusion and Future Directions}
This survey comprehensively synthesized feature matching methods across diverse modalities, including RGB images~\cite{liu2020crnet,liu2022few,yan2025subjective}, depth (RGB-D), LiDAR, 3D point clouds, medical imaging, and vision-language tasks. A prominent trend observed is the shift from traditional handcrafted techniques such as SIFT and SURF towards advanced deep learning approaches, which offer enhanced accuracy and adaptability. Additionally, detector-free architectures, exemplified by transformer-based dense matchers such as LoFTR, signify an important evolution towards robust, unified matching pipelines.

Feature matching methods have effectively adapted to the unique challenges posed by each modality. For instance, techniques designed for depth and LiDAR data emphasize geometric invariance, whereas solutions for medical imaging primarily address intensity variations. Despite these advancements, cross-modal matching continues to present significant challenges due to substantial representational disparities among different modalities. Addressing these requires either specialized cross-modal strategies or the development of truly modality-agnostic representations.

This survey provides extensive coverage, clearly differentiating handcrafted from learned methods and systematically comparing detector-based versus detector-free strategies. Additionally, it offers detailed insights into relevant datasets, evaluation protocols, and practical applications. Nevertheless, it acknowledges potential limitations, including possible gaps in benchmarks and challenges in keeping pace with the rapidly evolving landscape of deep learning methodologies. 

Robust feature matching remains fundamental to various computer vision tasks, including 3D reconstruction, simultaneous localization and mapping (SLAM), and object recognition. Despite deep learning-driven improvements, persistent challenges include limited generalization across diverse domains, computational inefficiencies restricting real-time applicability, and inherent complexities of cross-modal feature matching.

Looking ahead, the field is at an exciting juncture with promising opportunities to bridge traditionally isolated modalities within cohesive frameworks. Future research should prioritize developing adaptable, multi-modal pipelines that integrate RGB images, depth data, LiDAR scans, 3D point clouds, medical imaging, and vision-language modalities into unified systems.

Key future directions include, but are not limited to: developing modality-agnostic representations that bridge diverse sensor types and imaging conditions; designing computationally efficient, context-adaptive models for real-time and mobile deployment; establishing comprehensive benchmarks tailored to multi-modal, real-world scenarios; leveraging large foundation and generative models for versatile cross-modal matching; building unified multi-modal, multi-task frameworks that handle retrieval, registration, and captioning within a single system; and enabling lifelong learning for continual adaptation to new modalities and tasks without catastrophic forgetting. An extended discussion of each direction is provided in the supplementary material.

\section*{Acknowledgment}
This work was supported in part by the Ganpo Elite Talent Program under Grant GPYC20250030, in part by the National Natural Science Foundation of China under Grant 62476048, and in part by the Agency for Science, Technology and Research (A*STAR), Singapore, under the MTC Programmatic Funds (Grant M23L7b0021). This work was also supported in part by A*STAR, Singapore, under the MTC Programmatic Funding Scheme (Grant M23L8b0049) through the Scent Digitalization and Computation (SDC) Programme.

%% file: 8_Appendix.tex
\section*{Appendix A:Cross-Modality Vision-Language Feature Matching: From Traditional Methods to Large-Scale Foundation Models}

\noindent\textbf{Embodied AI and Vision-Language Navigation}\\
Embodied AI refers to AI agents that interact with environments, often in a physical or simulated 3D space. One of the flagship tasks in this category is \textbf{Vision-Language Navigation (VLN)}, where an agent is placed in an environment and must follow a natural language instruction (e.g., “Go down the hall and turn left into the kitchen, then stop by the refrigerator”) to reach a goal location.
The VLN task was popularized by Anderson et al., who introduced the Room-to-Room (R2R) dataset \cite{Anderson2018-nav}. R2R provided 7,189 paths in simulated houses (Matterport3D environments \cite{Chang2017}) with corresponding English instructions. The agent’s objective is to navigate from a start to an end location purely by following the instruction. 

Initial approaches to VLN used a sequence-to-sequence model with attention: an LSTM to encode the instruction, another LSTM to decode actions, attending to the instruction context at each step \cite{Anderson2018-nav}. 
One key difficulty in VLN is the mismatch between how instructions are given and how agents are trained. Early models struggled with generalization: they overfit to training environments, partly due to limited data. Research addressed this with data augmentation and better learning methods: \textbf{Speaker-Follower Models} introduced an idea of synthetic instruction-path pairs \cite{fried2018speaker}. \textbf{RCM} added an auxiliary progress monitor and used rewards for moving closer to the goal described by the instruction \cite{Wang2019RCM}. RL helps because it exposes the agent to its own mistakes during training and teaches it to correct course, whereas pure supervised learning can be brittle if anything goes off the reference path. Ma \etal added a self-monitoring component where the agent predicts how much of the instruction has been completed at each step \cite{Ma2019SelfMonitor}. 
More recently, \textbf{HAMT} \cite{chen2021history} uses transformers and attention over history, effectively remembering where the agent has been and what was seen, similar to memory in navigation.

Beyond R2R, many variants of the task were created: \textbf{Room-Across-Room (RxR)} provided multilingual instructions (English, Hindi, Telugu) and longer instructions per path, increasing diversity and requiring models to handle multiple languages \cite{Ku2020RxR}.
\textbf{Touchdown} addressed outdoor navigation in Google Street View with very long instructions and a final "touchdown" location to pinpoint \cite{Chen2019Touchdown}. 
\textbf{Vision-and-Dialog Navigation (CVDN)} combined navigation with a dialog agent, where the agent could ask questions when confused \cite{Thomason2020CVDN}. This involves a “guide” and a “follower” having a dialog, bringing language interaction into the loop.
\textbf{REVERIE} was a task that combined VLN with object grounding: the instruction refers to an object that is not explicitly named by category \cite{Qi2020REVERIE}. The agent must navigate to the correct room and also identify the referent object. This bridges navigation with referring expression resolution.
\textbf{ALFRED} took embodiment further by adding interaction: ALFRED is an indoor instructional task where the agent has to not just navigate but also manipulate objects to complete a task (like “put a potato in a microwave and turn it on”) \cite{Shridhar2020ALFRED}. This requires vision, language, and action planning, including picking up objects, using receptacles, etc., in a simulated environment (AI2-THOR simulator \cite{Kolve2017}).

Embodied vision-language tasks have revealed some unique challenges not present in static tasks: The need to handle partial observability (the agent only sees what’s in front of it; it must remember or explore for unseen info).
Longer-term planning: instructions can be long, and subgoals might need to be inferred.
Error compounding: A single missed instruction step can lead the agent far off course.
Also, sim-to-real gap: ultimately, we want agents that can do this in the real world. Issues like different visuals, continuous motion, and safety come into play. 

Despite these challenges, embodied AI with language grounding is making strides. The combination of computer vision for perception, NLP for understanding and generation, and reinforcement learning for decision making makes it one of the most interdisciplinary and rich areas of AI. It is likely to benefit further from the trend of foundation models: e.g., a large language model could be used to better interpret instructions or to handle dialogues (by providing common-sense reasoning), while a vision model like Grounded SAM~\cite{ren2024grounded}, which combines text-conditioned grounding transformers with the Segment Anything Model (SAM), could help with open-world object detection and manipulation from natural language instructions.

\begin{table*}[!htbp]\small
\centering
\small
\caption{The vision-language datasets with their typical use-cases, content description, and evaluation protocols. These benchmarks have driven the development of methods discussed in this survey.}
\label{tab:datasets-cross-vision-language}

\begin{tabular}{p{1.1cm} p{1.6cm} p{4.8cm} p{5.2cm}}
\toprule
\textbf{Dataset} & \textbf{Use-case} & \textbf{Description} & \textbf{Evaluation Protocol} \\
\midrule
GQA \cite{Hudson2019} & Compositional VQA & 113K images (VG) with 22M synthesized questions that require multi-step reasoning (balanced to reduce language bias). Annotations include structured scene graphs for the images. & Accuracy on open-ended answers. Also provides consistency and validity metrics (consistency: answer similar questions similarly; validity: answer is in a plausible range). Usually evaluated on a balanced test split with $\sim$3M questions. \\
CLEVR \cite{Johnson2017} & Diagnostic VQA & 100K synthetic images of 3D shapes and 1M generated questions testing compositional reasoning (counting, comparing attributes, logic). & Accuracy (exact match) on each question type. Because the dataset is balanced and free of bias, overall accuracy reflects reasoning ability. Human performance is nearly 100\%; model performance indicates specific reasoning failures. \\
Visual Dialog \cite{das2017visual} & Image-grounded Dialogue & 123k dialogs (QA sequences) on COCO images. Each dialog has 10 question-answer pairs, where each question is asked based on the image and conversation history. & Two modes: retrieval (choose the correct answer from 100 candidates at each turn, evaluated by mean reciprocal rank, Recall@1) and generative (free-form answer generation, evaluated by BLEU or human judgment). Also, track dialog consistency and use of image evidence qualitatively. \\
Room-to-Room (R2R) \cite{Anderson2018-nav} & Vision-Language Navigation & 21.5k instructions for 7,189 paths in Matterport3D houses. Each instruction averages 29 words describing a route through connected panoramic views. & Success Rate (SR): fraction of trajectories where the agent's end location is within 3m of the goal. SPL: SR normalized by path length (penalizes longer-than-necessary routes). Also, Oracle's success (if any visited location was the goal). Evaluated on unseen houses in the val/test. \\
RxR \cite{Ku2020RxR} & Multilingual VLN & 126k instructions (in English, Hindi, Telugu) for 12k paths in Matterport3D. Richer, more diverse instructions, some with multiple sentences. & Same metrics as R2R (SR, SPL), reported per-language as well. Tests cross-lingual generalization (training often on English and testing on other languages). \\
REVERIE \cite{Qi2020REVERIE} & VLN + Referring & Uses Matterport3D scenes: 10k navigation instructions that refer to remote objects (not visible from start). Combining navigation to a room and then identifying a referred object by a short phrase. & Two metrics: Navigation success and Grounding success. “Success on Remote Vision-and-Dialog” is measured by both navigation and identification being correct. \\
\bottomrule
\end{tabular}
\end{table*}

\noindent\textbf{Visual Question Answering (VQA)} \\
VQA was introduced as a grand challenge for joint vision and language understanding by Antol et al., who released the VQA v1 dataset with open-ended questions about MSCOCO images \cite{vqa15}. \\
Early models followed an encoder–fusion–decoder paradigm: image and question features are encoded and then fused (e.g., concatenation plus MLP) to predict answers \cite{vqa15}. Stacked Attention Networks (SAN) refined this by applying one or multiple rounds of question-guided attention over image features \cite{Yang2016}, and co-attention models further introduced joint attention over both image regions and question words \cite{Lu2016}. 
Fusion mechanisms were then strengthened. Multimodal Compact Bilinear pooling used an approximated outer product to capture higher-order interactions between image and text features \cite{fukui2016multimodal}, while BAN employed Bilinear Attention Networks that coupled attention with bilinear fusion, achieving strong performance \cite{kim2018bilinear}. \\
VQA v2 \cite{goyal2017making} was later proposed to mitigate dataset biases in v1 \cite{vqa15} via balanced image pairs, encouraging models to rely more on visual evidence. Region-based features from Bottom-Up and Top-Down Attention \cite{Anderson2018-bottom} (using Faster R-CNN object proposals instead of CNN grids) further improved visual grounding. 
Beyond standard VQA, related benchmarks include Visual Dialog \cite{das2017visual}, where the model answers a sequence of questions about an image while maintaining dialog context, and Visual Commonsense Reasoning (VCR) \cite{zellers2019recognition}, which requires answering questions that involve commonsense reasoning and inference beyond the image.

\section*{Appendix B: Modality-Specific Overview Diagrams}

\begin{figure}[!htbp]
    \centering
    \includegraphics[width=\textwidth]{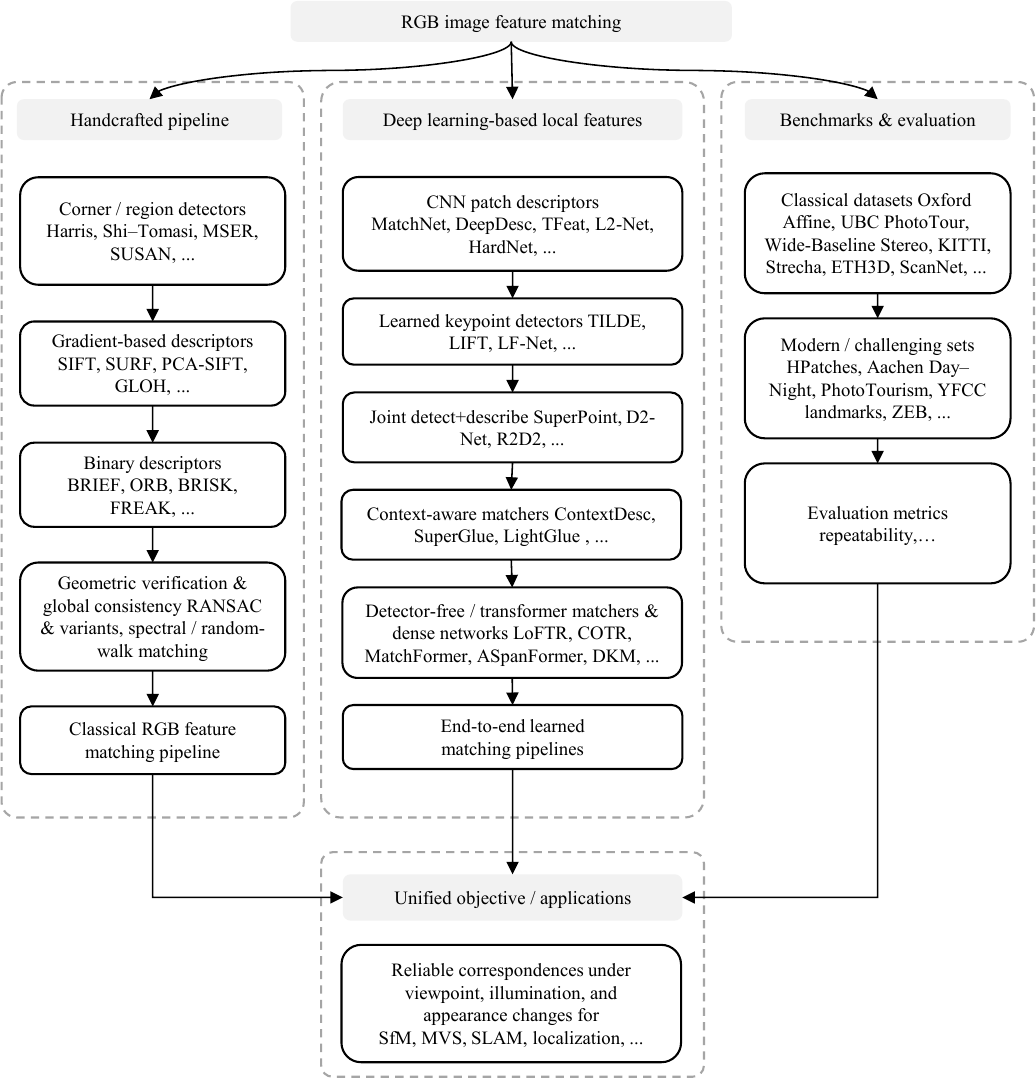}
    \caption{Overview of RGB image feature matching. The diagram summarizes the classical pipeline 
    from corner/region detection and local descriptor extraction to geometric verification, alongside 
    modern deep learning--based local features (learned detectors, joint detect--describe networks, 
    and dense/transformer matchers) and standard benchmarks.}
    \label{fig:rgb-matching-overview}
\end{figure}

\begin{figure}[!htbp]
    \centering
    \includegraphics[width=0.8\textwidth]{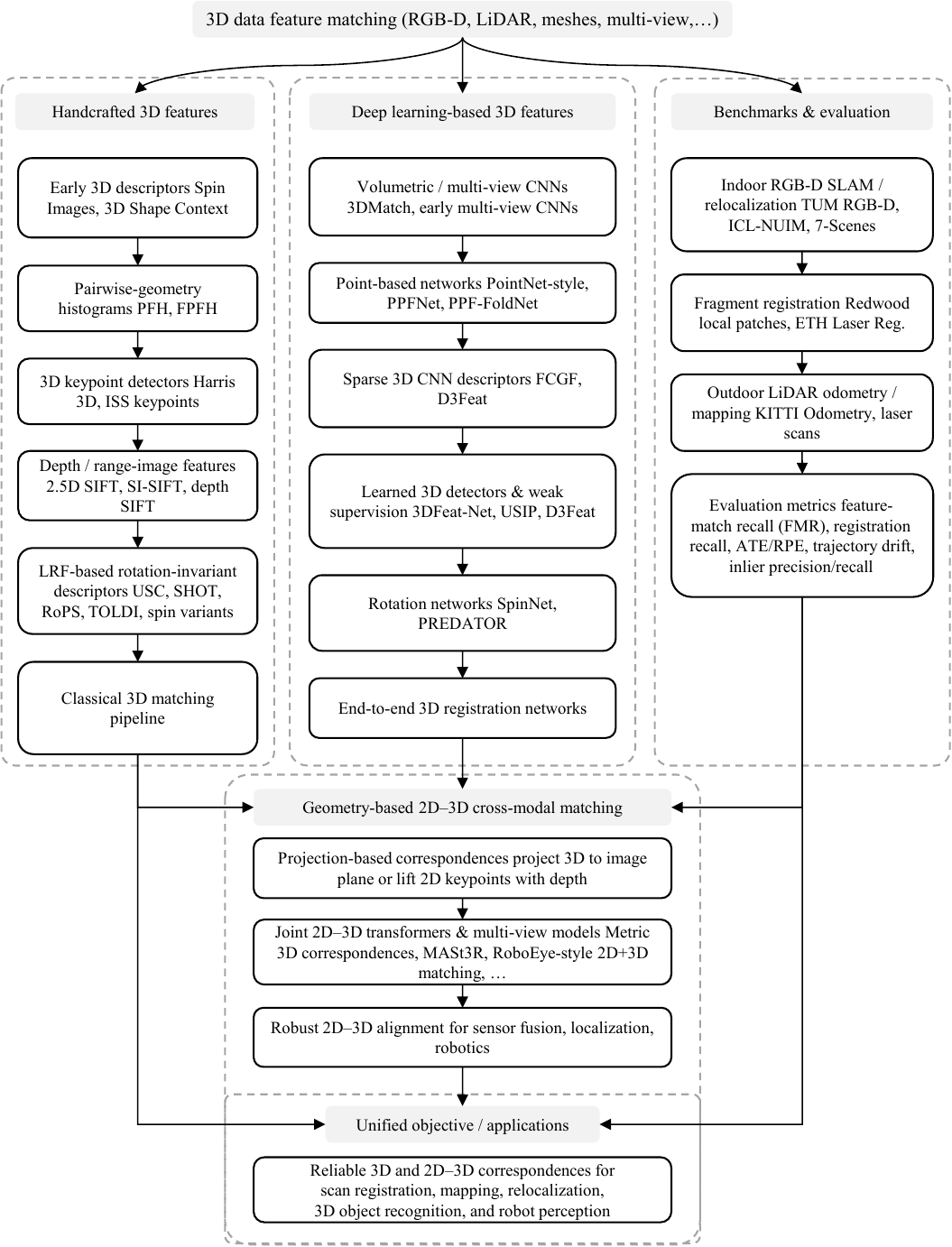}
    \caption{Overview of 3D data feature matching for RGB--D, LiDAR, meshes, and multi-view 3D. 
    The figure contrasts handcrafted 3D features (early descriptors, pairwise-geometry histograms, 
    3D keypoint detectors, and LRF-based rotation-invariant descriptors) with deep learning--based 
    approaches (volumetric and multi-view CNNs, point-based networks, sparse 3D CNN descriptors, 
    learned detectors, rotation/overlap-aware networks, and end-to-end registration). It also 
    highlights geometry-based 2D--3D cross-modal matching with joint 2D--3D transformers, standard 
    benchmarks, and evaluation metrics.}
    \label{fig:3d-matching-overview}
\end{figure}

\begin{figure}[!htbp]
    \centering
    \includegraphics[width=0.9\textwidth]{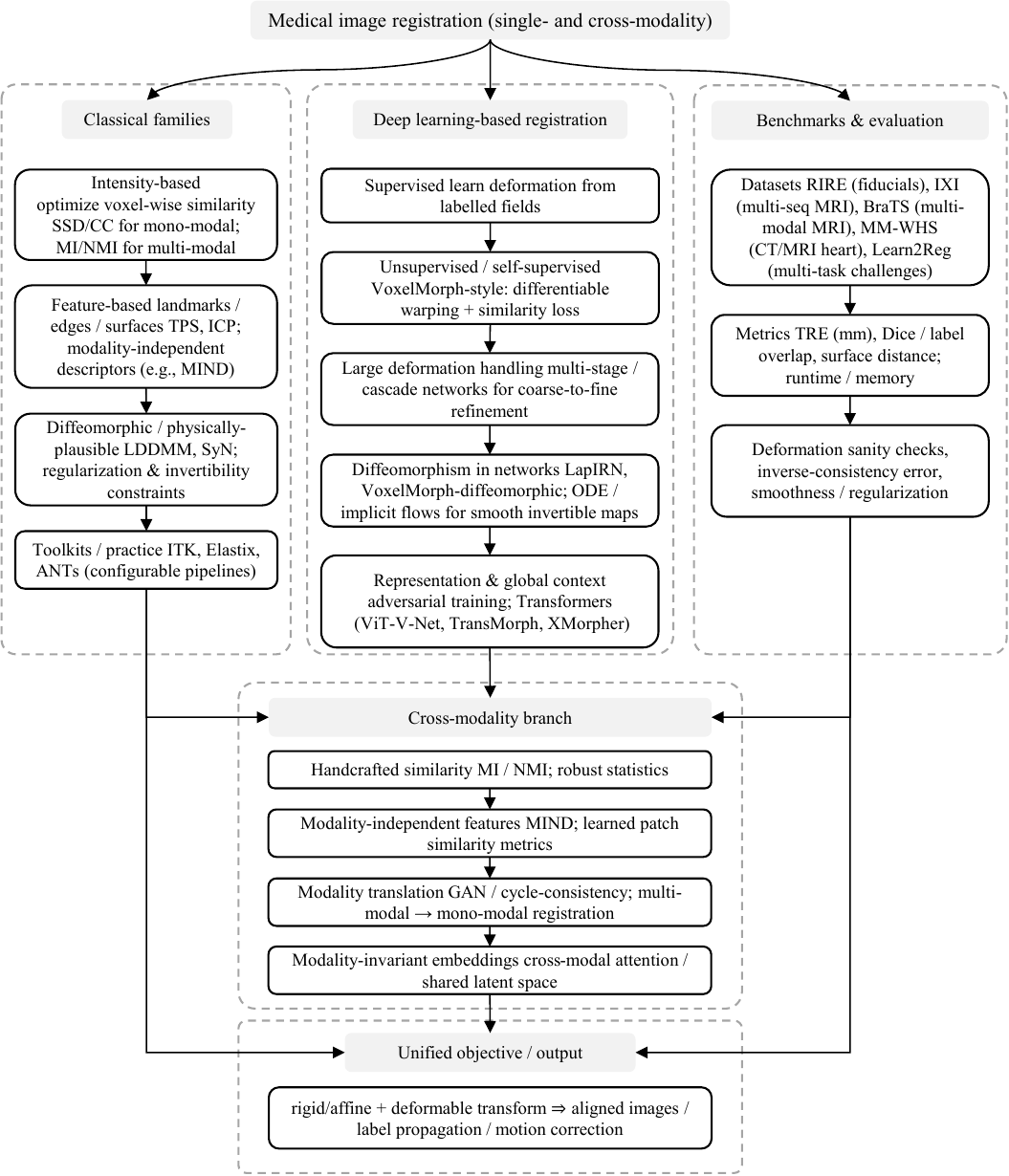}
    \caption{Modality-aware view of medical image registration across single- and cross-modality 
    settings. The left branch summarizes classical families, including intensity-based methods, 
    feature-based registration with landmarks and modality-independent descriptors, diffeomorphic 
    formulations, and widely used toolkits. The middle branch covers deep learning--based registration, 
    from supervised and unsupervised VoxelMorph-style networks to cascade/large-deformation models, 
    diffeomorphic neural flows, and transformer-based architectures. The right branch focuses on 
    cross-modality strategies such as handcrafted similarity measures, modality-independent features, 
    image-to-image translation, and modality-invariant embeddings. Representative datasets and 
    evaluation metrics are listed, while the unified objective is to estimate rigid/affine and 
    deformable transforms for accurate alignment, label propagation, and motion correction.}
    \label{fig:medical-registration-overview}
\end{figure}

\begin{figure}[!htbp]
    \centering
    \includegraphics[width=\textwidth]{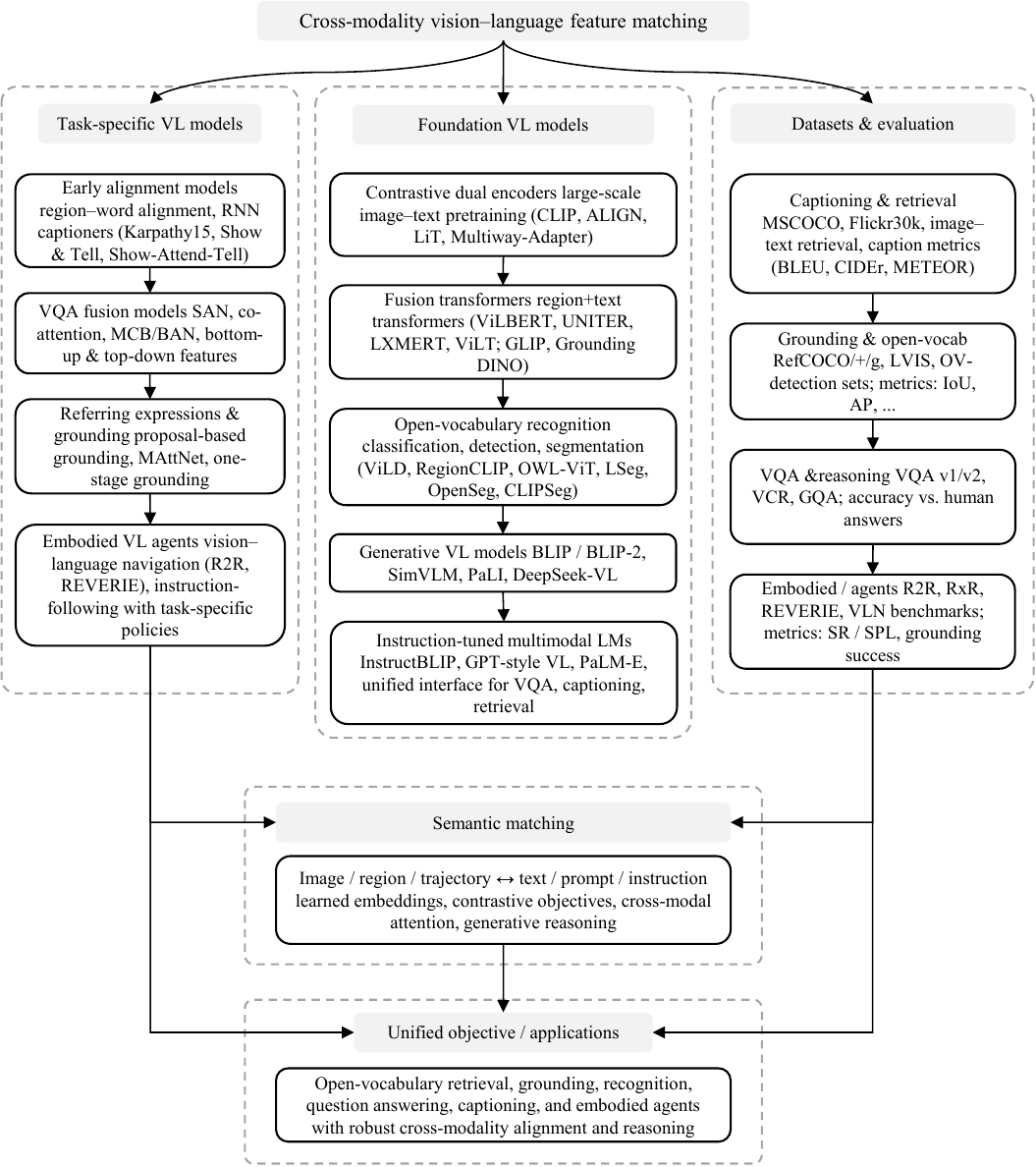}
    \caption{Cross-modality vision--language feature matching viewed as semantic correspondence 
    between visual content (images, regions, trajectories) and text (captions, prompts, instructions). 
    The figure organizes task-specific vision--language models (early alignment and captioning models, 
    VQA fusion networks, referring expression grounding, and embodied VL agents) and foundation 
    vision--language models (contrastive dual encoders, fusion transformers, open-vocabulary 
    recognition models, generative VL models, and instruction-tuned multimodal LMs), together with 
    typical datasets and evaluation metrics. }
    \label{fig:vl-matching-overview}
\end{figure}

For completeness, this appendix provides four modality-specific overview diagrams that visually
summarize the main pipelines and design spaces discussed in the survey. 
Figure~\ref{fig:rgb-matching-overview} presents the landscape of RGB image feature matching, 
illustrating the evolution from classical detector–descriptor pipelines to modern dense and 
transformer-based matchers. 
Figure~\ref{fig:3d-matching-overview} extends this perspective to 3D data, including RGB--D, LiDAR, 
meshes, and multi-view 3D, highlighting both handcrafted 3D descriptors and deep learning approaches. 
Figure~\ref{fig:medical-registration-overview} focuses on medical image registration, organizing 
classical, deep learning, and cross-modality strategies within a unified modality-aware view. 
Finally, Figure~\ref{fig:vl-matching-overview} summarizes cross-modality vision--language feature 
matching, emphasizing semantic correspondences between visual content and text across a range of 
tasks and foundation models.

\section*{Appendix C: Future Directions}
Key future directions include, but are not limited to:
\begin{itemize}
    \item  \textbf{Modality-Agnostic Representations:} Developing robust, generalized representations that effectively bridge diverse sensor types and imaging conditions. Leveraging advances in self-supervised learning, foundation models, and transformer architectures will be crucial for achieving robust representations and reducing modality-aware engineering overhead.

    \item  \textbf{Computational Efficiency:} Emphasizing the design of lightweight and resource-efficient models that adapt dynamically to context and available modalities. Optimizing network structures and integrating hardware acceleration will significantly enhance real-time and mobile application deployment.

    \item  \textbf{Comprehensive Benchmarking:} Establishing realistic and extensive benchmarking protocols tailored specifically to multi-modal and real-world scenarios. These benchmarks will provide critical guidance to align research efforts with practical application needs and to promote the development of integrated, robust, and universally applicable feature matching solutions.

    \item \textbf{
    Foundation and Generative Models for Multi-Modal Matching:} The rise of large foundation models offers a path toward versatile cross-modal feature matching. Models like CLIP and BLIP already learn alignments between vision and language. Extending this paradigm, future work can develop foundation models that jointly encode RGB images, 3D scans, medical images, and more into a shared representational space. 
    Generative models like diffusion networks also hold potential: their cross-modal capabilities (e.g., text-to-image generation) and rich internal representations can be repurposed for feature matching tasks. For instance, the attention maps of a text-to-image diffusion model have been used to align semantic regions across modalities. By leveraging large-scale pretraining and billion-parameter models, these approaches aim for high generalizability: a single model could be tuned for diverse matching tasks with minimal modality-aware engineering.

    \item \textbf{Unified Multi-Modal and Multi-Task Frameworks:} An ambitious yet increasingly tangible goal is a unified architecture capable of handling multiple modalities and tasks within one system. Instead of maintaining separate pipelines for RGB vs. depth vs. text, or for retrieval vs. registration vs. captioning, a single framework could flexibly accommodate all. 
    Such a unified model can simultaneously perform image–text retrieval, 3D scene alignment, medical image registration, or even captioning and question-answering, by simply switching inputs and prompts. The benefits of unification include efficiency and seamless modality interplay. 
    The trend toward "all-in-one" models is growing, and we anticipate systems that can ingest anything from natural images to point clouds to radiology scans, and produce whichever output is required, all under a cohesive framework.

    \item \textbf{Lifelong Learning and Continual Adaptation:} Finally, future feature matching systems are expected to learn continuously, adapting to new modalities and tasks over time without forgetting past knowledge. Lifelong learning would enable a deployed matching system to self-improve as it encounters novel conditions. For example, a robot’s matching module could acquire new skills when a new sensor (thermal camera, ultrasound, etc.) is added, or a medical image registration network could continually update itself as it sees new types of scans.
    Research may exploit strategies like experience replay, modular expansion, or meta-learning to allow models to evolve. Few-shot learning will be crucial while the aim is to incorporate drastically different modalities or tasks using only a small number of new examples. 

\end{itemize}